\documentclass[sigconf,screen]{acmart}
\usepackage{algorithm}
\usepackage{algorithmic}
\usepackage{bbding}

\usepackage{multicol,multirow}
\usepackage{bbding}
\usepackage{comment}
\usepackage{bm}
\usepackage{balance}

\newtheorem{assumption}{Assumption}[section]

\DeclareMathOperator*{\argmax}{arg\,max}

\newcommand{\mE}{\mathbb{E}}
\newcommand{\mV}{\mathbb{V}}
\newcommand{\calD}{\mathcal{D}}
\newcommand{\calX}{\mathcal{X}}
\newcommand{\calA}{\mathcal{A}}
\newcommand{\calE}{\mathcal{E}}

\newcommand{\calN}{\mathcal{N}}
\newcommand{\calZ}{\mathcal{Z}}

\newcommand{\calS}{\mathcal{S}}

\newcommand{\mx}{\bm{x}}
\newcommand{\ma}{\bm{a}}
\newcommand{\ms}{\bm{s}}
\newcommand{\me}{\bm{e}}

\newcommand{\ips}{\hat{V}_{\mathrm{IPS}} (\pi; \calD)}

\newcommand{\ours}{\hat{V}_{\mathrm{LIPS}} (\pi; \calD)}

\newcommand{\pinv}{\hat{V}_{\mathrm{PI}} (\pi; \calD)}

\definecolor{dkred}{rgb}{0.8,0,0}
\definecolor{dkpink}{rgb}{1.0,0,0.5}
\definecolor{tickgreen}{rgb}{0,0.6,0}

\newcommand{\mse}{\mathrm{MSE} ( \hat{V} (\pi ) )}
\newcommand{\bias}{\mathrm{Bias} ( \hat{V} (\pi; \calD) )}
\newcommand{\var}{\mV_{\calD} \big[ \hat{V} (\pi; \calD ) \big]}
\usepackage{multirow}

\AtBeginDocument{%
  \providecommand\BibTeX{{%
    \normalfont B\kern-0.5em{\scshape i\kern-0.25em b}\kern-0.8em\TeX}}}

\copyrightyear{2024}
\acmYear{2024}
\setcopyright{acmlicensed}\acmConference[WWW '24]{Proceedings of the ACM Web Conference 2024}{May 13--17, 2024}{Singapore, Singapore}
\acmBooktitle{Proceedings of the ACM Web Conference 2024 (WWW '24), May 13--17, 2024, Singapore, Singapore}
\acmDOI{10.1145/3589334.3645343}
\acmISBN{979-8-4007-0171-9/24/05}

\author{Haruka Kiyohara}
\authornote{This work was done during their internship at CyberAgent, Inc.}
\affiliation{Cornell University}
\email{hk844@cornell.edu}

\author{Masahiro Nomura}
\affiliation{CyberAgent, Inc.}
\email{nomura\_masahiro@cyberagent.co.jp}

\author{Yuta Saito}
\affiliation{Cornell University}
\email{ys552@cornell.edu}

\begin{document}
\title[Off-Policy Evaluation of Slate Bandit Policies via Optimizing Abstraction]{Off-Policy Evaluation of Slate Bandit Policies\\via Optimizing Abstraction}


\renewcommand{\shortauthors}{Haruka Kiyohara et al.}

\begin{abstract}
We study \textit{off-policy evaluation} (OPE) in the problem of slate contextual bandits where a policy selects multi-dimensional actions known as slates. This problem is widespread in recommender systems, search engines, marketing, to medical applications, however, the typical Inverse Propensity Scoring (IPS) estimator suffers from substantial variance due to large action spaces, making effective OPE a significant challenge. The PseudoInverse (PI) estimator has been introduced to mitigate the variance issue by assuming linearity in the reward function, but this can result in significant bias as this assumption is hard-to-verify from observed data and is often substantially violated. To address the limitations of previous estimators, we develop a novel estimator for OPE of slate bandits, called \textit{Latent IPS} (LIPS), which defines importance weights in a low-dimensional slate abstraction space where we optimize slate abstractions to minimize the bias and variance of LIPS in a data-driven way. By doing so, LIPS can substantially reduce the variance of IPS without imposing restrictive assumptions on the reward function structure like linearity. Through empirical evaluation, we demonstrate that LIPS substantially outperforms existing estimators, particularly in scenarios with non-linear rewards and large slate spaces.
\end{abstract}



\maketitle
\section{Introduction}
Slate bandits play a pivotal role in many online services, such as recommender and advertising systems. In these services, a decision-making algorithm or \textit{policy} selects a combinatorial and potentially high-dimensional slate composed of multiple sub-actions. For example, visual advertisements include elements like the title, key visual, and background image, each of which significantly impacts user engagement and revenue. Similarly, in medical treatment scenarios, selecting the optimal combination of drug doses to enhance patient outcomes is of paramount importance. While these systems generally have access to extensive logged data, accurately estimating the effectiveness of a new slate bandit policy through Off-Policy Evaluation (OPE) presents significant challenges. These challenges are due to the exponential increase in variance associated with slate-wise importance weighting needed in the slate bandit setup~\citep{swaminathan2017off,su2020doubly}.

Despite the challenges in OPE with slate structures, it remains a crucial task for safely evaluating and improving the effectiveness of real-world interactive systems~\citep{saito2021counterfactual}. Therefore, several \textit{PseudoInverse} (PI) estimators have been proposed~\citep{su2020doubly, swaminathan2017off, vlassis2021control}. These estimators use slot-wise importance weights to reduce the variance of IPS and have been shown to enable unbiased OPE under the \textit{linearity} assumption on the reward. This assumption requires that the expected reward should be linearly decomposable, ignoring interaction effects among different slots. While PI often outperforms IPS under linear reward structures, it can still produce significant bias when linearity does not hold. In addition, the variance of PI remains extremely high if there are many unique sub-actions~\citep{saito2022off}. It is worth noting that while several estimators have been developed for OPE in ranking action spaces~\citep{kiyohara2022doubly,kiyohara2023off,li2018offline,mcinerney2020counterfactual}, they are not applicable to evaluate slate bandits. This is because these estimators assume that the rewards are available for every position in a ranking, whereas in our setup, only slate-wise rewards are available.

To overcome the limitations of slate OPE, we propose a novel approach called \textit{Latent IPS} (LIPS), which redefines importance weights in a low-dimensional slate abstraction space. LIPS is inspired by recent advances in OPE for large action spaces where the marginalized IPS (MIPS) estimator leverages pre-defined action embeddings to provably reduce variance~\citep{saito2022off,saito2023off}, but our estimator does not assume that action embeddings are already observed in the logged data a priori as MIPS. Through theoretical analysis, we show that LIPS can substantially reduce variance compared to IPS while also being unbiased if the slate abstraction is \textit{sufficient} in that it retains sufficient information to characterize the reward function. In addition, LIPS can achieve lower bias and variance than PI under appropriate slate abstractions since our sufficiency condition is more relaxed than the linearity assumption and the variance of LIPS only depends on the size of the abstraction space, which can be more compact than the sub-action spaces. Interestingly, our analysis on the bias and variance of LIPS also implies that its mean-squared-error (MSE) may be minimized when strategically using an insufficient slate abstraction, potentially resulting in even greater variance reduction while being nearly unbiased. Based on this theoretical analysis, we also develop a procedure to optimize slate abstraction to directly minimize the bias and variance of LIPS. This is a particular distinction of LIPS from MIPS~\citep{saito2022off} and its extension~\citep{saito2023off}, which assume that useful action embeddings already exist. Empirical results on extreme classification datasets (which can be transformed into bandit feedback with slate actions) demonstrate that LIPS enables more accurate slate OPE than IPS, PI, a naive extension of MIPS, and their doubly-robust variants on a variety of non-linear reward functions and large slate action spaces. 

Our contributions can be summarized below.
\begin{itemize}
    \item We propose the LIPS estimator, which leverages an abstraction of slates to substantially improve OPE of slate bandits.
    \item We develop a fully data-driven procedure to optimize an abstraction function to directly minimize the MSE of LIPS.
    \item We empirically demonstrate that LIPS with an optimized abstraction outperforms existing estimators (such as PI, IPS, and MIPS) for a range of scenarios with non-linear rewards.
\end{itemize}

\section{Off-Policy Evaluation for Slates}
This work considers a \textit{slate} contextual bandit problem where $\mx \in \calX \subseteq \mathbb{R}^d$ represents a context vector (e.g., user demographics, consumption history, weather) and $\ms \in \calS := \prod_{l=1}^L \calA_l$ denotes a slate action. A slate consists of several sub-actions, i.e., $\ms = (a_1, a_2, \ldots, a_L)$ where each sub-action $a_l$ is chosen from the corresponding action set $\calA_l$, which may differ across different slots. For instance, in an email campaign, $\calA_1$ may be a set of subject lines, while $\calA_2$ may represent whether or not to include visuals in a promotion email.

We refer to a function $\pi: \calX \rightarrow \Delta(\calS)$ as a slate bandit policy, which maps each context to a distribution over the slate space. In particular, throughout the paper, we will focus on a \textit{factored} policy, i.e., $\pi(\ms\,|\,\mx) = \prod_{l=1}^L \pi(a_l\,|\,\mx)$ for brevity of exposition. Let then $r$ be a reward associated with a slate $\ms$, which is considered sampled from an unknown conditional distribution $p(r \,|\, \mx, \ms)$. We are interested in OPE in this slate bandit setup, where we are given a logged dataset $\calD := \{ (\mx_i, \ms_i, r_i) \}_{i=1}^n$ collected by a logging policy $\pi_0$ where $(\mx, \ms, r) \sim p(\mx)\pi_0(\ms|\mx)p(r|\mx,\ms)$. In particular, we aim to estimate the following \textit{value} of a new policy $\pi$ (which is called a target policy and is often different from $\pi_0$):
\begin{align*}
    V(\pi)
    &:= \mE_{(\mx, \ms) \sim p(\mx)\pi(\ms|\mx)}[q(\mx, \ms)],
\end{align*}
where $q(\mx, \ms) := \mE[r \,|\, \mx, \ms]$ is the expected reward function given context $\mx$ and slate $\ms$. In particular, our goal is to develop an estimator $\hat{V}$ capable of accurately estimating the value of $\pi$ relying only on the logged data $\calD$. The accuracy of an estimator is typically measured by the mean-squared-error (MSE): 
\begin{align*}
    \mse :&= \mE_{\calD} \big[ \big(  V(\pi) - \hat{V} (\pi; \calD) \big)^2 \big] \\
    &= \bias^2 + \var,
\end{align*}
where the bias and variance of $\hat{V}$ are defined respectively as
\begin{align*}
    \bias &:= \mE_{\calD}[\hat{V} (\pi; \calD)] - V(\pi), \\
    \var &:= \mE_{\calD} \big[ \big(  \hat{V} (\pi; \calD) - \mE_{\calD} [\hat{V} (\pi; \calD)] \big)^2 \big].
\end{align*}

\paragraph{\textbf{Existing estimators.}} 
We now summarize the key existing estimators and their limitations.

\paragraph{Inverse Propensity Scoring (IPS)~\citep{strehl2010learning}}
IPS reweighs the observed rewards by the ratios of slate probabilities under the target and logging policies (slate-wise importance weight) as
\begin{align}
    \ips
    := \frac{1}{n} \sum_{i=1}^n \biggl( \prod_{l=1}^L \frac{\pi(a_{i,l}\,|\,\mx_i)}{\pi_0(a_{i,l}\,|\,\mx_i)} \biggr) r_i = \frac{1}{n} \sum_{i=1}^n w(\mx_i,\ms_i) r_i,
\end{align}
where $w(\mx,\ms) := \pi(\ms|\mx)/\pi_0(\ms|\mx)$ is the slate-wise importance weight. IPS is unbiased under some identification assumptions such as common support (i.e., $\pi(\ms\,|\,\mx) > 0 \rightarrow \pi_0(\ms\,|\,\mx) > 0,\,\forall (\mx, \ms)$). However, its critical issue is its exponential variance, which arises due to the potentially astronomical size of the slate action space~\citep{swaminathan2017off}.

\paragraph{PseudoInverse (PI)~\citep{swaminathan2017off}} 
To deal with the exponential variance of IPS, the PI estimator leverages only the slot-wise importance weights (compared to slate-wise importance weighting of IPS) as
\begin{align}
    \pinv := \frac{1}{n} \sum_{i=1}^{n} \left(\sum_{l=1}^L \frac{\pi\left(a_{i, l} \mid \mx_{i}\right)}{\pi_0\left(a_{i, l} \mid \mx_{i}\right)}-L+1\right) r_{i}.
\end{align}
Since PI relies only on slot-wise importance weights, its variance only scales with the number of unique sub-actions (i.e., $\sum_{l=1}^L |\calA_l|$). Thus, the variance of PI is often smaller than that of IPS whose weight scales with the cardinality of the slate space (i.e., $\prod_{l=1}^L |\calA_l|$). PI has also been shown to enable an unbiased performance evaluation, i.e, $\mE_{\calD} [ \pinv ] = V(\pi)$, under the linearity assumption, which requires that the reward function is linearly decomposable and there exists some (latent) intrinsic reward functions $\{\phi_l\}_{l=1}^L$ such that $q(\mx, \ms) = \sum_{l=1}^L \phi_l(\mx, a_l)$ for every context $x$ and slot $l$. In other words, this assumption essentially ignores every possible interactions across different slots.

Although PI improves the MSE over IPS under linear reward functions, its bias is no longer controllable when linearity does not hold, which is often the case in highly non-linear real-world environments. In addition, PI may still suffer from extremely high variance when $\sum_{l=1}^L |\calA_l|$ is large (i.e., when there are many unique sub-actions). These limitations of PI and IPS motivate the development of a new estimator for slate OPE that can substantially reduce the variance while being (nearly) unbiased without any unrealistic assumptions on the reward function structure.

\section{The LIPS Estimator}
To deal with the issues of IPS and PI, we now propose the LIPS estimator that enables more effective OPE by leveraging \textit{slate abstraction} rather than positing restrictive assumptions on the reward. At a high level, LIPS defines importance weights in a (low-dimensional) latent slate space $\calZ$, which can either be discrete or continuous\footnote{In the remainder of the paper, we rely on a discrete abstraction for ease of exposition, but a continuous abstraction space can also be considered under stochastic abstraction.} and is induced via a slate abstraction function $\phi_{\theta}: \mathcal{S} \rightarrow \mathcal{Z}$, parametrized by $\theta$. Specifically, our LIPS estimator is defined as
\begin{align}
    \ours 
    := \frac{1}{n} \sum_{i=1}^n \underbrace{\frac{\pi(\phi_{\theta}(\ms_i) \,|\, \mx_i)}{\pi_0(\phi_{\theta}(\ms_i) \,|\, \mx_i)}}_{:= w(\mx_i, \phi_{\theta}(\ms_i))} r_i \label{eq:lips}
\end{align}
where $\pi(z | \mx) := \sum_{\ms \in\calS} \mathbb{I}\{ \phi_{\theta}(\ms) = z \} \pi(\ms| \mx)$ is a marginal distribution of an abstracted slate induced by policy $\pi$, and $w(\mx,z) := \pi(z | \mx) / \pi_0(z | \mx)$ is the \textit{latent importance weight}. Note that we can readily extend LIPS to the case with a context-dependent and stochastic abstraction based on a parameterized distribution $p_{\theta}: \mathcal{X} \times \mathcal{S} \rightarrow \Delta(\mathcal{Z})$. This means that we can generalize Eq.~\eqref{eq:lips} as
\begin{align*}
    \ours 
    := \frac{1}{n} \sum_{i=1}^n \frac{p_{\theta}(z_i \,|\, \mx_i;\pi)}{p_{\theta}(z_i \,|\, \mx_i;\pi_0)} r_i
\end{align*}
where $z_i \sim p_{\theta}(z\,|\,\mx_i, \ms_i)$ and $p_{\theta}(z \,|\,\mx;\pi) := \sum_{\ms \in \calS} \pi(\ms\,|\,\mx) p_{\theta}(z\,|\,\mx, \ms)$ $= \sum_{\ms \in \calS} p_{\theta}(\ms,z\,|\,\mx;\pi)$. This extension allows for a flexible control of the bias-variance tradeoff of LIPS and ensures a tractable optimization of abstraction. The central idea of LIPS is to circumvent the reliance on slate- or slot-wise importance weights, substantially improving the variance from IPS and PI while avoiding the linear-reward assumption like PI. The following formally analyzes LIPS and shows its statistical advantages over the existing estimators. We also develop a data-driven optimization procedure for slate abstractions to directly minimize the MSE of LIPS.

\subsection{Theoretical Analysis} \label{sec:theoretical}
First, we analyze the bias of LIPS based on the notion of \textit{sufficient} slate abstraction (but we will later show that intentionally using an insufficient slate abstraction is indeed a better implementation and present how to data-drivenly obtain a better slate abstraction.)

\begin{definition} (Sufficient Slate Abstraction) \label{def:sufficient}
A slate abstraction function $\phi_{\theta}$ is said to be \textit{sufficient} if it satisfies $q(\mx,\ms)=q(\mx,\ms')$ for all $\mx\in\calX$, $\ms\in\calS$, and $\ms' \in \{\ms'' \in \calS \mid \phi_{\theta}(\ms) = \phi_{\theta}(\ms'')\}$.
\end{definition}

\noindent An abstraction function is \textit{sufficient} if it aggregates only the slates that have the same expected reward, and it means that the latent slate space $\calZ$ retains sufficient information to characterize the reward function. For example, identity abstraction $\phi(\ms)=\ms$ is always sufficient (LIPS is reduced to IPS in this case). Note here that this notion of sufficiency does not impose any particular restriction on the reward function form such as linearity. Furthermore, sufficient slate abstractions may not be unique, and there could potentially be many sufficient slate abstractions.

The following demonstrates that LIPS is unbiased when given a sufficient abstraction function. We also characterize the bias of LIPS when we use a stochastic abstraction, which may not be sufficient.

\begin{theorem} (Unbiasedness of LIPS) \label{thrm:unbiased}
LIPS is unbiased, i.e., 
$$\mE_{\calD}[\ours] = V(\pi),$$ if a given slate abstraction function $\phi_{\theta}$ is sufficient. See Appendix~\ref{app:unbiased} for the proof.
\end{theorem}

\begin{theorem} (Bias of LIPS) \label{thrm:bias}
The bias of LIPS given a stochastic slate abstraction $p_{\theta}$ is
\begin{align}
    & \mathrm{Bias(\ours)} \\
    &= \mE_{p(\mx)p_{\theta}(z|\mx;\pi_0)} \bigl[ \sum_{j < k \leq |\calS|} p_{\theta}(\ms_j\,|\,\mx,z;\pi_0) p_{\theta}(\ms_k\,|\,\mx,z;\pi_0) \notag \\
    & \quad \quad \quad \quad \quad \quad \times (q(\mx,\ms_j) - q(\mx,\ms_k)) \times (w(\mx,\ms_k) - w(\mx,\ms_j)) \bigr] \notag,
\end{align}
where $p_{\theta}(\ms\,|\,\mx,z;\pi)= p_{\theta}(\ms,z\,|\,\mx;\pi)/\pi(z\,|\,\mx)$. See Appendix~\ref{app:bias} for the proof.
\end{theorem}

\noindent In particular, Theorem~\ref{thrm:bias} implies that the bias of LIPS is characterized by the following factors.
\begin{enumerate}
    \item identifiability of the slates from their abstractions: \\
    $p_{\theta}(\ms_j|\mx,z;\pi_0) p_{\theta}(\ms_k|\mx,z;\pi_0)$
    \item difference in the expected rewards between a pair of slates: $q(\mx,\ms_j) - q(\mx,\ms_k)$
    \item difference in the slate-wise importance weights between a pair of slates: $w(\mx,\ms_k) - w(\mx,\ms_j)$
\end{enumerate}

\noindent When the slate $\ms$ is almost deterministic given $(\mx, z)$, the value of $p_{\theta}(\ms_j|\mx,z;\pi_0)$ tends to be either zero or one. As a result, the product $p_{\theta}(\ms_j|\mx,z;\pi_0) p_{\theta}(\ms_k|\mx,z;\pi_0)$ becomes nearly zero. This indicates that if the latent slate space sufficiently encapsulates the information required to reconstruct the original slates, then even with a stochastic slate abstraction, the bias of LIPS remains minimal. Furthermore, the second factor relates to the predictiveness of rewards given the latent variable $z$. Specifically, if the latent variable is \textit{almost sufficient}—meaning that slates with similar expected rewards tend to have similar abstraction distributions—the difference $q(\mx,\ms_j) - q(\mx,\ms_k)$ will be small, leading to a reduced bias of LIPS. Thus, the analysis implies that the bias of LIPS remains small if the latent slate space is finer-grained and abstraction is closer to deterministic, which makes the latent variable $z$ more predictive of slate $\ms$ and reward $r$.

Next, we analyze the variance reduction of LIPS against IPS, which can be substantially large depending on the coarseness and stochasticity of the latent slate space.

\begin{theorem} (Variance Reduction) \label{thrm:variance}
Given a sufficient slate abstraction function $\phi_{\theta}$, we have
\begin{align}
    &n \bigl( \mV_{\calD} [\ips ] - \mV_{\calD} [\ours ] \bigr) \notag \\
    &= \mE_{p(\mx)\pi_0(\phi_{\theta}(\ms)|\mx)} \left[ \mE_{p(r|\mx,\phi_{\theta}(\ms))} [ r^2 ] \,\mV_{\pi_0(\ms|\mx,\phi_{\theta}(\ms))} \left[ w(\mx,\ms) \right]  \right].
\end{align}
See Appendix~\ref{app:variance-reduction} for the proof.
\end{theorem}

\begin{table*}[t]
    \centering
    \caption{A toy example illustrating the potential advantage of strategic variance reduction with an insufficient abstraction. LIPS with an insufficient (but optimized) abstraction produces much smaller variance while introducing some small bias, resulting in a smaller MSE than LIPS with a sufficient abstraction.}
    \label{tab:strategic-variance-reduction}
    \vspace{-2mm}
    \scalebox{1.1}{
    \begin{tabular}{c|ccc}
        \toprule
         & bias & variance & \textbf{MSE} (= bias$^2$ + variance) \\ \midrule
        LIPS with a sufficient abstraction & 0.0 & 1.0 & \textbf{1.00} ($\,= (0.0)^2 + 1.0$)\\
        LIPS with an insufficient (but optimized) abstraction & 0.5 & 0.2 &  \textbf{0.45} ($\,=(0.5)^2 + 0.2$)\\
        \bottomrule
    \end{tabular}}
\end{table*}

\noindent There are two primary factors that determine the degree of variance reduction. The first is the second moment of the reward, which increases when the reward has high variability. The second is the variance of the slate-wise importance weight $w(\mx,\ms)$ regarding the conditional distribution $\pi_0(\ms|\mx,\phi(\ms))(=\pi_0(\ms|\mx)/\pi_0(\phi(\ms)|\mx))$. This factor escalates when \textbf{(i)} $w(\mx,\ms)$ attains large values and \textbf{(ii)} $\pi_0(\ms|x,\phi(\ms))$ maintains sufficient stochasticity. This suggests that variance reduction of LIPS could be exponential, as the variance of the slate importance weight may increase exponentially with the slate size $L$. Additionally, the degree of variance reduction can be modulated through the cardinality of the latent slate space and the entropy of the slate abstraction. In other words, a coarser and stochastic abstraction leads to greater variance reduction. A key theoretical insight here is that a sufficient abstraction might not always minimize the MSE of LIPS. Table~\ref{tab:strategic-variance-reduction} presents a hypothetical example demonstrating how LIPS with an insufficient abstraction may achieve a lower MSE than with a sufficient abstraction. Specifically, LIPS with a sufficient abstraction achieves the MSE of 1.0 with zero bias as per Theorem~\ref{thrm:unbiased}. However, a lower MSE can be realized by intentionally using an insufficient abstraction. This approach allows for a substantial reduction in variance (-0.8) at the cost of a minor increase in squared bias (+0.25). Thus, a sufficient abstraction does not necessarily yield the optimal MSE for LIPS. Consequently, rather than seeking a sufficient abstraction, the following sections describe a data-driven strategy to \textit{optimize} the abstraction to minimize the MSE of LIPS.

\subsection{Optimizing Slate Abstractions} \label{sec:abstraction}
The analysis from the previous section indicates that \textbf{the bias-variance tradeoff of LIPS is determined by the granularity of the latent slate space $\calZ$ and stochasticity of slate abstraction distribution $p_{\theta}(z \,|\,\mx,\ms)$}, which also suggests that the MSE of LIPS might be minimized with an \textit{insufficient} abstraction that yields even greater variance reduction while producing only some small bias. This insight naturally encourages us to directly minimize the bias and variance of LIPS when optimizing slate abstraction rather than myopically searching for a sufficient abstraction. Specifically, we propose to optimize slate abstraction distribution $p_{\theta}$ via
\begin{align}
    & (\hat{\theta}, \hat{\psi}, \hat{\omega}) = \argmax_{\theta,\psi}\min_{\omega} \sum_{i=1}^n \mathcal{L}(\mx_i, \ms_i, \pi_0; \theta,\psi,\omega)
\end{align}
where
\begin{align}
    & \mathcal{L}(\mx, \ms, \pi_0; \theta,\psi,\omega) 
    &= \underbrace{\mE_{p_{\theta}(z\,|\,\mx, \ms;\pi_0)}\bigl[\mathrm{log} \, p_{\psi}(\ms\,|\,\mx,z;\pi_0)\bigr]}_{\text{bias reduction}: \,\,\, p_{\theta}(\ms_j\,|\,\mx,z;\pi_0)p_{\theta}(\ms_k\,|\,\mx,z;\pi_0)} \\
    & & \quad + \underbrace{\mE_{p_{\theta}(z\,|\,\mx, \ms;\pi_0)}\bigl[ (r - \hat{q}_{\omega}(\mx, z))^2 \bigr]}_{\text{bias reduction}: \,\,\, q(\mx,\ms_j) - q(\mx, \ms_k)} \notag \\
    & & \quad \underbrace{- \beta \mathrm{KL}(p_{\theta}(z\,|\,\mx,\ms;\pi_0) \,||\, p_{\psi}(z\,|\,\mx;\pi_0))}_{\text{variance reduction}: \,\,\, p_{\theta}(\ms\,|\,\mx,z;\pi_0)}.
\end{align}
$\theta$, $\psi$, and $\omega$ are the parameters of slate abstraction $p_{\theta}(z\,|\,\mx, \ms;\pi_0)$, slate reconstruction $p_{\psi}(\ms\,|\,\mx,z;\pi_0)$, and reward construction $\hat{q}_{\omega}(\mx,z)$ models.\footnote{The pseudo-code (Algorithm~\ref{algo:abstraction}) of this abstraction optimization procedure can be found in the appendix. Appendix~\ref{app:proof} provides a more extensive bias-variance analysis of LIPS when given a stochastic abstraction $p_{\theta}$. Note that, if some action features are available, it is also possible to use such action features concatenated across slates to the models to further improve the abstraction optimization procedure.} The first two terms of the loss function aim for bias reduction, while the last term controls variance reduction of LIPS. More specifically, the first term measures the identifiability of the slates and the second term measures how predictive the rewards are based on the latent variable. The final term acts as a regulator for the stochasticity of $p_{\theta}(\ms\,|\,\mx,z;\pi_0)$ by aligning $p_{\theta}(z\,|\,\mx,\ms;\pi_0)$ more closely with $p_{\psi}(z\,|\,\mx;\pi_0)$. This will indeed regularize the latent importance weights and make them close to $1$ everywhere (as described in the appendix in detail), and thus we can expect a larger variance reduction of LIPS when the last term is dominant. The regularization weight $\beta$ is thus considered the key hyper-parameter that governs the bias-variance tradeoff of LIPS (i.e., a smaller $\beta$ implies a smaller bias and larger variance of LIPS, while a larger $\beta$ leads to a larger bias and smaller variance). This hyper-parameter can be tuned by existing parameter tuning methods such as SLOPE~\citep{su2020adaptive} and PAS-IF~\citep{udagawa2023policy}, which are feasible using only observed logged data $\calD$. Section~\ref{sec:experiment} empirically investigates how LIPS works with these existing parameter tuning methods.

\section{Related Work}

\paragraph{\textbf{Off-Policy Evaluation}}
Off-Policy Evaluation (OPE) has gained increasing attention in fields ranging from recommender systems to personalized medicine as a safe alternative to online A/B tests, which might be risky, slow, and sometimes even unethical~\citep{saito2022counterfactual,kiyohara2021accelerating,kiyohara2023scope,kiyohara2024towards}. Among many OPE estimators studied in the single-action setting, DM and IPS~\citep{strehl2010learning} are commonly considered baseline estimators. DM trains a reward prediction model to estimate the policy value. Although DM does not produce large variance, it can be highly biased when the reward predictor is inaccurate. In contrast, IPS allows for unbiased estimation under standard identification assumptions, but it often suffers from high variance due to large importance weights. Doubly Robust (DR)~\citep{dudik2014doubly} is a hybrid method that combines DM and IPS to improve variance while remaining unbiased. However, its variance can still be very high under large action spaces~\citep{saito2022off}. As a result, the primary objective of OPE research has been to effectively balance the bias and variance, and numerous estimators have been proposed to address this statistical challenge~\citep{metelli2021subgaussian,su2020doubly,wang2016optimal,saito2021evaluating}. 

In comparison, the slate contextual bandit setting has been relatively under-explored despite its prevalence in real practice~\citep{dimakopoulou2019marginal,ie2019slateq,swaminathan2017off} and the necessity for significant variance reduction due to combinatorial action spaces. Existing approaches, such as PI~\citep{swaminathan2017off} and its variants~\citep{su2020doubly, vlassis2021control}, strongly rely on the linearity assumption of the reward function. However, when this assumption does not hold, PIs are no longer unbiased. Moreover, their variance can be substantial when there are many unique sub-actions. Compared to PIs, LIPS improves the variance without making restrictive assumptions about the reward function form via leveraging slate abstraction. LIPS can have a much lower bias and variance than PI when slate abstraction is appropriately optimized. Our approach is also relevant to the MIPS~\citep{saito2022off} and OffCEM~\citep{saito2023off} estimators, which employ action embeddings in the single-action setting. Although they assume that useful action embeddings are already present in the logged data, we develop a novel method to optimize slate abstractions based on the logged bandit data to directly improve the resulting estimator. The next section empirically demonstrates that our data-driven optimization procedure to obtain an appropriate slate abstraction is a crucial component of LIPS to outperform the existing estimators (IPS and PI) as well as MIPS~\cite{saito2022off}.

It is important to note that there exists a relevant line of research called OPE of ranking policies, where the action space consists of ordered sets of items~\citep{kiyohara2022doubly,li2018offline,mcinerney2020counterfactual,kiyohara2023off}. While this ranking setting closely resembles that of the slate, all existing estimators require the observation of slot-wise rewards, which makes them inapplicable to our slate setup, where only the reward for each slate is observed (thus the slate OPE problem is fundamentally more difficult). However, the idea of abstraction could be effective in OPE for ranking policies as well, and we consider this to be a valuable future direction.

\paragraph{\textbf{Abstraction in Bandits}}
Action abstraction is often used in bandits to accelerate policy learning through efficient exploration~\citep{martin2022nested,moraes2018action,sen2021top,slivkins2011multi}. For example, \citep{martin2022nested, slivkins2011multi} propose a tree-based hierarchical structure for the action space, where each tree node represents an action abstraction, grouping similar actions in its child nodes. This structure effectively decreases the number of actions considered during exploration, thus enhancing sample efficiency in online policy learning. Similar concepts have been applied to improve the learning efficiency of ranking policies~\citep{sen2021top}. While these concepts inspire our approach, their primary focus is on policy learning. To our knowledge, no existing work has applied action abstraction to enhance OPE specifically for slate bandits.

\paragraph{\textbf{Representation Learning for Causal Inference and Recommendations}}
Latent variable modeling has been deemed effective in causal inference when identifying confounders from proximal variables~\citep{louizos2017causal} or recovering confounders and treatments from high-dimensional data such as texts~\citep{fong2021causal,veitch2020adapting,wood2018challenges}. Among them, the closest to ours is \citet{veitch2020adapting}, which takes into account reward prediction and treatment reconstruction loss when optimizing latent text representations to perform causal inference regarding text data. However, these methods have been developed for the conventional task of estimating the average treatment effect, and we are not aware of any similar approaches or loss functions in OPE.

In another line of research, latent action representations are often learned to exploit the structure in the action space in reinforcement learning~\citep{allshire2021laser,chandak2019learning,deffayet2023generative,zhao2019rethinking,zhou2021plas}. In particular, \citet{deffayet2023generative} use a Variational AutoEncoder (VAE) model to pre-train latent slate space from logged data to improve recommendations. Consequently, they achieve a better exploration-exploitation tradeoff regarding long-term objectives and item diversity through VAE. While the VAE model of \citep{deffayet2023generative} is somewhat similar to our optimization procedure for slate abstraction described in Section~\ref{sec:abstraction}, our loss function is derived from the theoretical analysis of LIPS and aims to directly improve the MSE in slate OPE rather than recommendation effectiveness.

\begin{figure*}[t]
\begin{minipage}[c]{0.99\hsize}
\centering
\scalebox{0.95}{
\begin{tabular}{c}
\begin{minipage}{0.80\hsize}
\begin{center}
\includegraphics[width=0.95\linewidth]{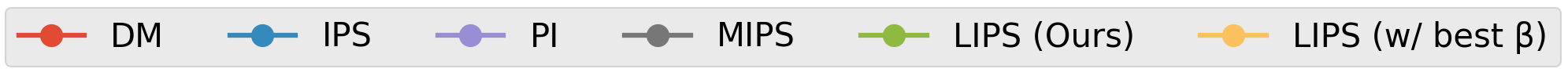}
\vspace{-2mm}
\end{center}
\end{minipage}
\\
\\
\begin{minipage}{0.99\hsize}
    \begin{center}
        \includegraphics[clip, width=0.9\linewidth]{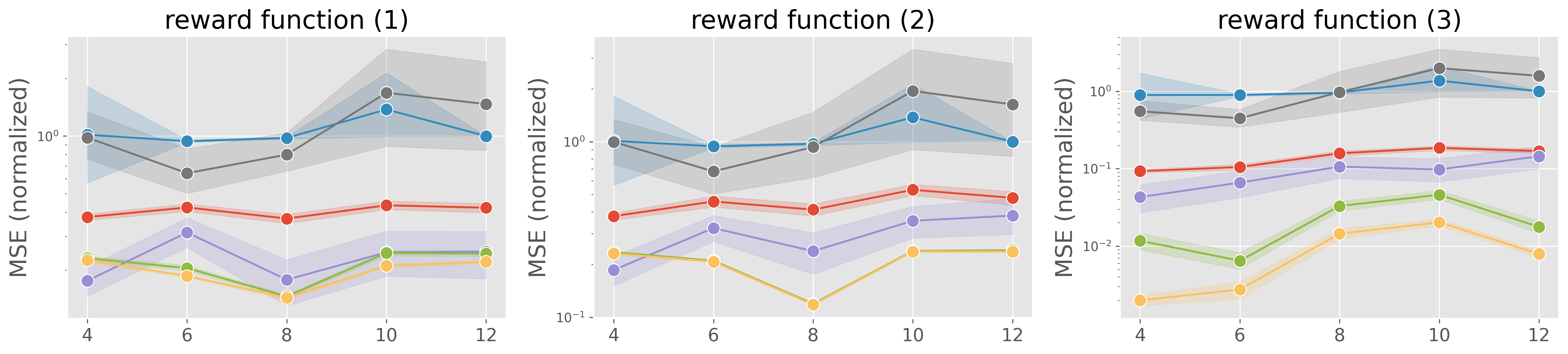}
        \vspace{-2mm}
        \caption{Comparison of the estimators' MSE (normalized by the true value $V(\pi)$) with varying slate sizes ($L$) and with reward functions (1) - (3) on the Wiki10-31K dataset. Note that the y-axis is on a log scale, thus the fluctuations observed for LIPS's MSE are less than 0.01.}
        \label{fig:slate_size_wiki}
        \vspace{3mm}
    \end{center}
\end{minipage}
\\ 
\\
\begin{minipage}{0.99\hsize}
    \begin{center}
        \includegraphics[clip, width=0.9\linewidth]{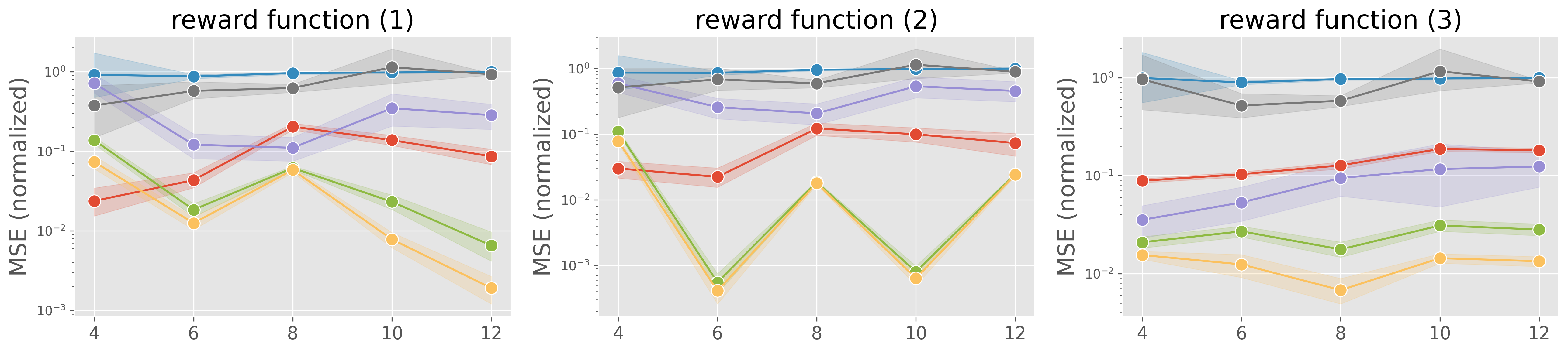}
        \vspace{-2mm}
        \caption{Comparison of the estimators' MSE (normalized by the true value $V(\pi)$) with varying slate sizes ($L$) and with reward functions (1) - (3) on the Eurlex-4K dataset. Note that the y-axis is on a log scale, thus the fluctuations observed for LIPS's MSE are less than 0.01.}
        \label{fig:slate_size_eurlex}
        \vspace{3mm}
    \end{center}
\end{minipage}
\\
\\
\end{tabular}
}
\end{minipage}
\end{figure*}
\begin{figure*}[t]
\begin{minipage}[c]{0.99\hsize}
\centering
\scalebox{0.95}{
\begin{tabular}{c}
\begin{minipage}{0.80\hsize}
\begin{center}
\includegraphics[width=0.95\linewidth]{figs/label.png}
\vspace{-2mm}
\end{center}
\end{minipage}
\\
\\
\begin{minipage}{0.99\hsize}
    \begin{center}
        \includegraphics[clip, width=0.9\linewidth]{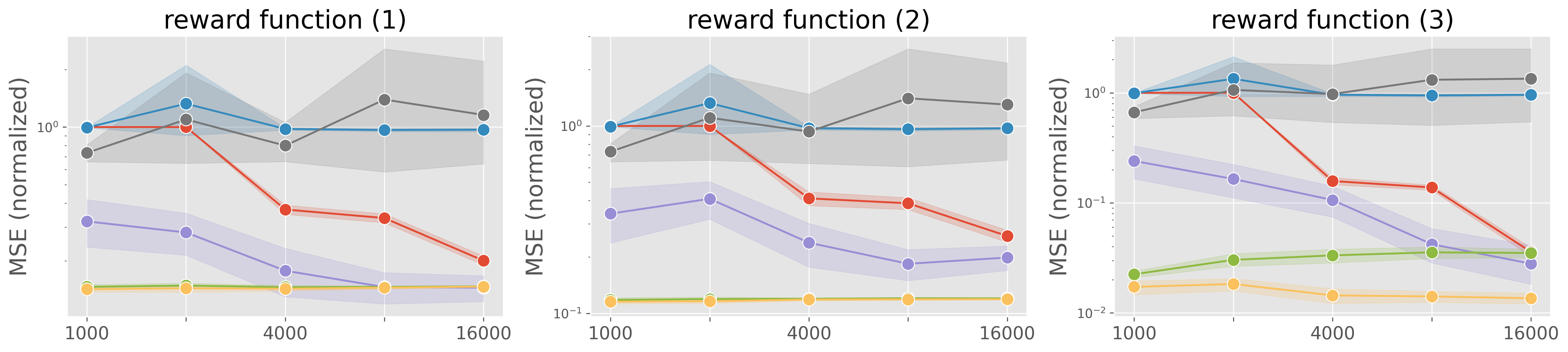}
        \vspace{-2mm}
        \caption{Comparison of the estimators' MSE (normalized by the true value $V(\pi)$) with varying data sizes ($n$) and with reward functions (1) - (3) on the Wiki10-31K dataset. Note that the y-axis is on a log scale.}
        \label{fig:data_size_wiki}
        \vspace{3mm}
    \end{center}
\end{minipage}
\\
\\
\begin{minipage}{0.99\hsize}
    \begin{center}
        \includegraphics[clip, width=0.9\linewidth]{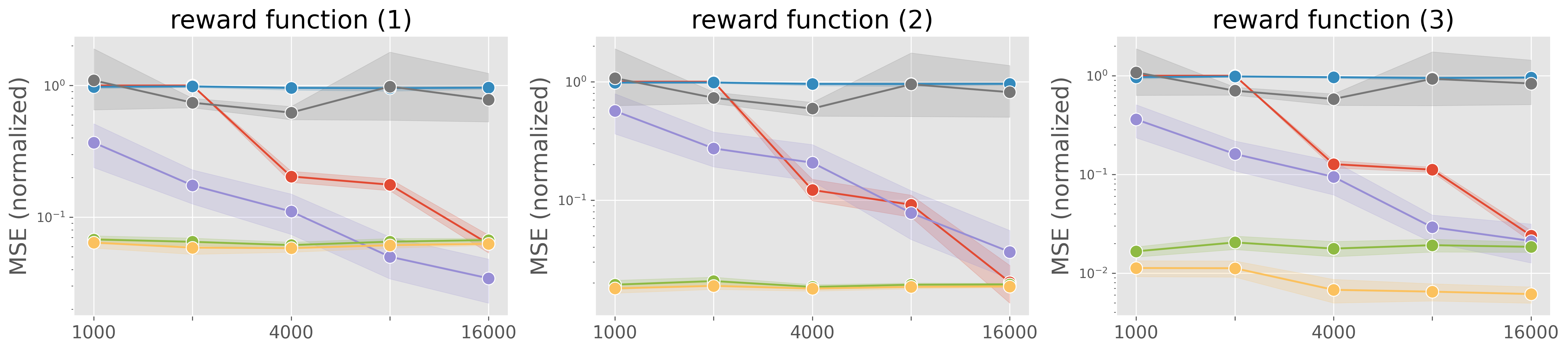}
        \vspace{-2mm}
        \caption{Comparison of the estimators' MSE (normalized by the true value $V(\pi)$) with varying data sizes ($n$) and with reward functions (1) - (3) on the Eurlex-4K dataset. Note that the y-axis is on a log scale.}
        \label{fig:data_size_eurlex}
        \vspace{3mm}
    \end{center}
\end{minipage}
\\
\\
\end{tabular}
}
\end{minipage}
\vspace{-4mm}
\end{figure*}
\section{Empirical Evaluation} \label{sec:experiment}
This section empirically compares LIPS with many relevant estimators on two real-world datasets. Our experiment code is available at \href{https://github.com/aiueola/webconf2024-slate-ope-via-abstraction}{\textcolor{dkpink}{https://github.com/aiueola/webconf2024-slate-ope-via-abstraction}}.

\subsection{Experiment Setup} 
We follow the standard "supervised-to-bandit" procedure to conduct an OPE experiment based on classification data, as used in many previous studies~\citep{farajtabar2018more,su2020doubly,su2019cab,wang2016optimal}. Specifically, we use the extreme classification datasets called Wiki10-31K and Eurlex-4K~\citep{bhatia16}. These datasets consist of many documents associated with a large number of labels, which are approximately 31K for Wiki10-31K and 4K for Eurlex-4K. Detailed dataset statistics are provided in Appendix~\ref{app:experiment}. 

To simulate a slate bandit scenario, we regard the documents as contexts ($\mx$). Wiki10-31K and Eurlex-4K inherently contain some text data that represent the documents, and we apply SentenceTransformer~\citep{reimers2019sentence} and PCA~\citep{abdi2010principal} to encode the raw texts into 20-dimensional contexts for both these datasets. Next, to define the slate action space, we first extract the top 1,000 dense labels in terms of the number of positive documents. Then, we randomly sample $L \times 10$ labels to form $L$ disjoint action sets $\{\calA_l\}_{l=1}^{L}$ corresponding to $L$ distinct slots, with each set having a size of $|\calA_l| = 10$. For each action $a \in \calA_l$, we first define $q_l(\mx, a) = 1 - \eta_a$ if the action has a positive label and $q_l(\mx, a) = \eta_a$ otherwise, where $\eta_a$ is a noise parameter sampled separately for each action from a uniform distribution with range $[0, 0.5]$. Given that we never know the nature of real-world reward functions within a slate, we simulate various \textit{non-linear} relationships (where linearity of PI is violated) that bring complex interactions across different slots. Specifically, we use the following synthetic reward functions (1) to (3).
\begin{align*}
    & \text{(1)}\, q(\mx, \ms) = \frac{1}{\lfloor L / 2 \rfloor} \sum_{l=1}^{\lfloor L / 2 \rfloor} q_l(\mx, a_l) + \frac{1}{\lfloor L / 2 \rfloor - 1}  \sum_{l=1}^{\lfloor L / 2 \rfloor -1} w(a_l, a_{l+1}) \\
    & \text{(2)}\,q(\mx, \ms) = \frac{1}{\lfloor L / 2 \rfloor} \Big( q_1(\mx, a_1) + \sum_{l=2}^{\lfloor L / 2 \rfloor} w(a_{l-1}, a_{l}) q_l(\mx, a_l) \Big) \\
    & \text{(3)}\,q(\mx, \ms) = \frac{1}{2} \Big( \min_{l=1}^{\lfloor L / 2 \rfloor} q_l(\mx, a_l) +  \max_{l=1}^{\lfloor L / 2 \rfloor} q_l(\mx, a_l) \Big)
\end{align*}

where $\lfloor c\rfloor := \max \{n \in \mathbb{Z} \mid n \leq c\}$ is the floor function. $w(a_l, a_{l+1})$ is a scalar value to represent the co-occurrence effect between $a_l$ and $a_{l+1}$ and it is sampled from the standard normal distribution. Note that, when defining these reward functions, we use only 50\% of the slots $l = 1, 2, \ldots, \lfloor L / 2 \rfloor$ among total of $L$ slots. By doing so, we can make $(a_1, a_2, \cdots, a_{\lfloor L / 2 \rfloor})$ a sufficient abstraction, which enables us to study if LIPS achieves more accurate estimation (lower MSE) by intentionally using an insufficient abstraction obtained by the optimization procedure from Section~\ref{sec:abstraction}.

Given these reward functions, we sample reward $r$ from a normal distribution as $r \sim \calN(q(\mx, \ms), \sigma^2)$ with a standard deviation of $\sigma = 0.1$. To obtain logging ($\pi_0$) and evaluation ($\pi$) policies, we first train a base classifier $\tilde{q}(\mx,\ms)$ via the REINFORCE method~\citep{williams1992simple} and then define the policies as follows:
\begin{align*}
    \pi_0(\ms \,|\, \mx) &= \prod_{l=1}^L \left( (1 - \epsilon_0) \frac{\exp(\gamma_0 \cdot \tilde{q}(\mx, a_l))}{\sum_{a \in \calA_l} \exp(\gamma_0 \cdot \tilde{q}(\mx, a))} + \frac{\epsilon_0}{|\calA_l|} \right), 
    \\
    \pi(\ms \,|\, \mx) &= \prod_{l=1}^L \left( (1 - \epsilon) \, \mathbb{I} \{ a_l = a_l^{\ast} \} + \frac{\epsilon}{|\calA_l|} \right),
\end{align*}
where $a_l^{\ast} := \argmax_{a \in \calA_l} \tilde{q}(\mx, a)$. $\gamma \in \mathbb{R}$ and $\epsilon_0, \epsilon \in [0, 1]$ are the experiment parameters that control the stochasticity of $\pi_0$ and $\pi$. We use $(\gamma,\epsilon_0, \epsilon)=(-1.0,0.1,0.3)$ in the main text.

\paragraph{\textbf{Compared estimators}}
We compare LIPS with the Direct Method (DM)~\citep{beygelzimer2009offset}, IPS~\citep{strehl2010learning}, PI~\citep{swaminathan2017off}, and MIPS~\citep{saito2022off}. DM is a regression-based estimator that estimates the policy value based on an estimated reward function $\hat{q}(\mx, \ms)$, which is learned by a neural network in our experiment. For LIPS, we employ a discrete abstraction whose dimension is 100 ($|\calZ| = 100$). When optimizing a slate abstraction, we select the hyper-parameter $\beta$ from $\{ 0.01, 0.1, 1.0, 10.0 \}$ based only on the available logged data via the SLOPE algorithm~\citep{su2020adaptive, tucker2021improved}. Appendix~\ref{app:experiment} provides some more details of SLOPE and describes how we parameterize and optimize the slate abstraction distribution $p_{\theta}$. Note that MIPS defines its importance weight taking only relevant slots (i.e., $l = 1, 2, \ldots, \lfloor L / 2 \rfloor$) into account and is defined as $\hat{V}_{\mathrm{MIPS}}(\pi;\calD) := \frac{1}{n} \sum_{i=1}^n \big( \prod_{l=1}^{\lfloor L/2 \rfloor} \frac{\pi(a_{i,l}\,|\,\mx_i)}{\pi_0(a_{i,l}\,|\,\mx_i)} \big) r_i,$ where the importance weights use only the first $\lfloor L/2 \rfloor$ slots as an action embedding leveraging the fact that the reward functions (1) - (3) depend only on these slots. MIPS is unbiased and has a lower variance than IPS, however, it is infeasible in practice since we do not know the true reward function. We include MIPS in our experiments since it is useful to investigate the effectiveness of intentionally using an insufficient abstraction. 
In addition to these baselines, we also report the results of ``LIPS (w/ best $\beta$)'' as a reference. It indicates LIPS with the best value of $\beta$ selected based on the ground-truth MSE, which provides the best accuracy achievable by our LIPS framework with an oracle hyper-parameter selection.

\begin{figure*}[t]
\begin{minipage}[c]{0.99\hsize}
\centering
\scalebox{0.95}{
\begin{tabular}{c}
\begin{minipage}{0.80\hsize}
\begin{center}
\includegraphics[width=1\linewidth]{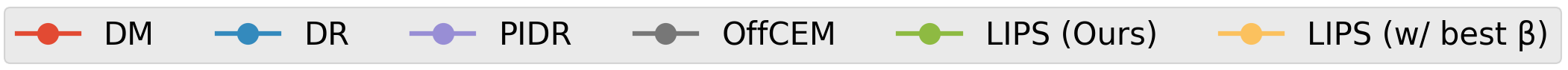}
\vspace{-2mm}
\end{center}
\end{minipage}
\\
\\
\begin{minipage}{0.99\hsize}
    \begin{center}
        \includegraphics[clip, width=0.95\linewidth]{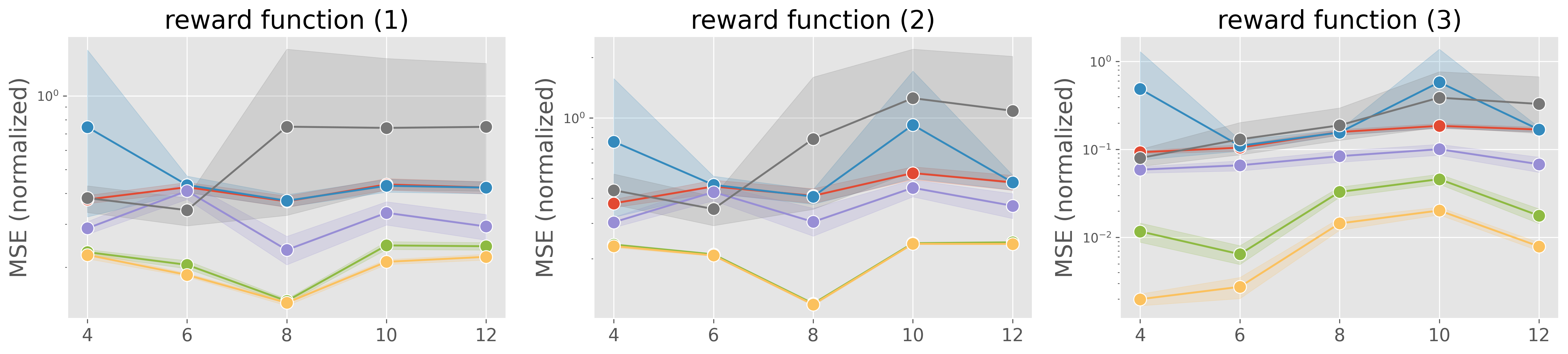}
        \vspace{-2mm}
        \caption{Comparison between LIPS and the DR estimators wrt their MSE (normalized by the true value $V(\pi)$) under varying slate sizes ($L$) and with reward functions (1) - (3) on the Wiki10-31K dataset. Note that the y-axis is on a log scale.}
        \label{fig:slate_size_wiki_dr}
        \vspace{3mm}
    \end{center}
\end{minipage}
\\ 
\\
\begin{minipage}{0.99\hsize}
    \begin{center}
        \includegraphics[clip, width=0.95\linewidth]{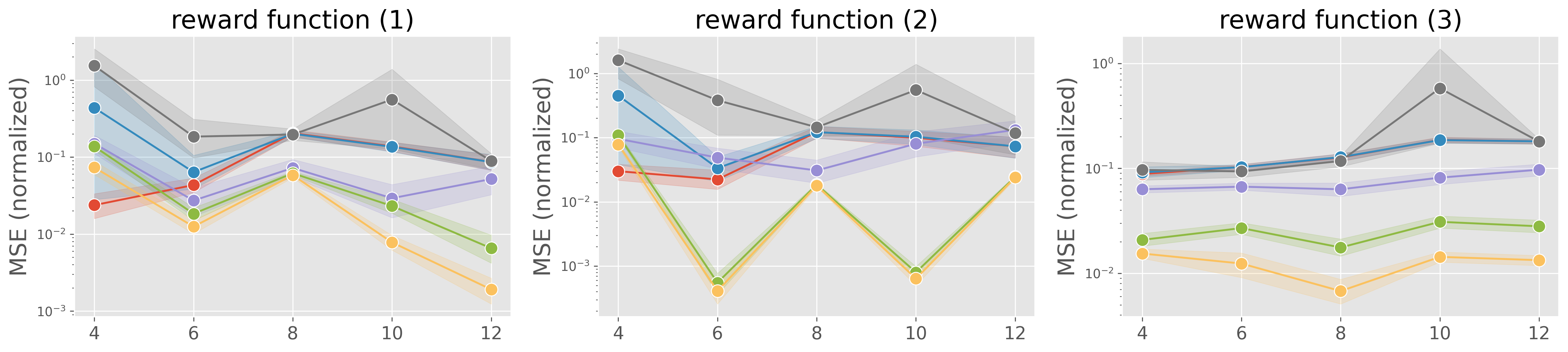}
        \vspace{-2mm}
        \caption{Comparison between LIPS and the DR estimators wrt their MSE (normalized by the true value $V(\pi)$) under varying slate sizes ($L$) and with reward functions (1) - (3) on the Eurlex-4K dataset. Note that the y-axis is on a log scale.}
        \label{fig:slate_size_eurlex_dr}
        \vspace{3mm}
    \end{center}
\end{minipage}
\\
\\
\begin{minipage}{0.99\hsize}
    \begin{center}
        \includegraphics[clip, width=0.95\linewidth]{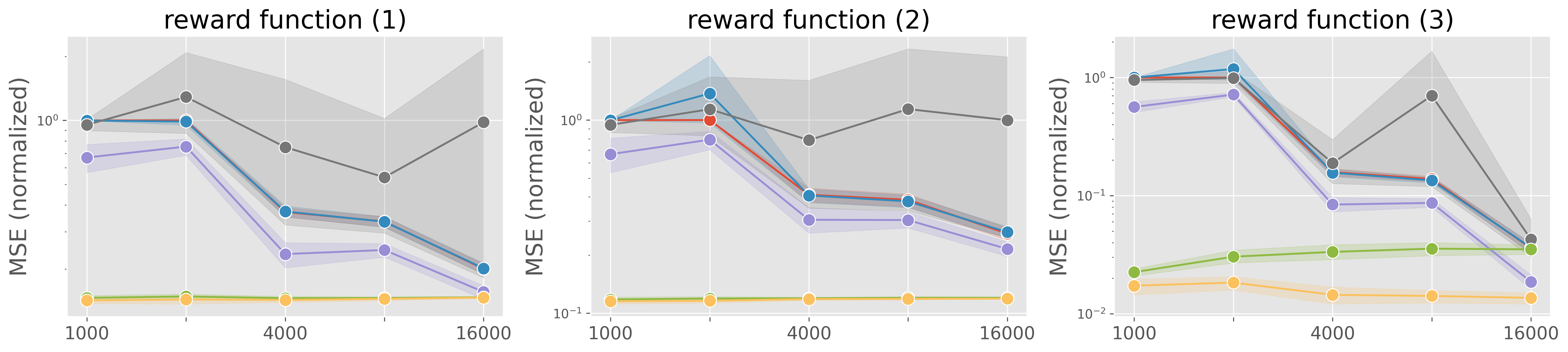}
        \vspace{-2mm}
        \caption{Comparison between LIPS and the DR estimators wrt their MSE (normalized by the true value $V(\pi)$) under varying data sizes ($n$) and with reward functions (1) - (3) on the Wiki10-31K dataset. Note that the y-axis is on a log scale.}
        \label{fig:data_size_wiki_dr}
        \vspace{3mm}
    \end{center}
\end{minipage}
\\ 
\\
\begin{minipage}{0.99\hsize}
    \begin{center}
        \includegraphics[clip, width=0.95\linewidth]{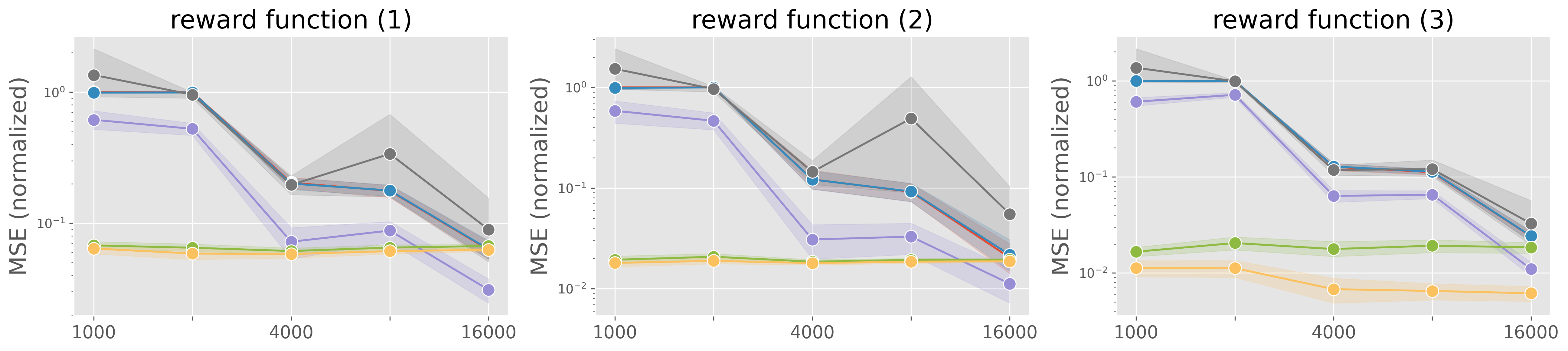}
        \vspace{-2mm}
        \caption{Comparison between LIPS and the DR estimators wrt their MSE (normalized by the true value $V(\pi)$) under varying data sizes ($n$) and with reward functions (1) - (3) on the Eurlex-4K dataset. Note that the y-axis is on a log scale.}
        \label{fig:data_size_eurlex_dr}
        \vspace{3mm}
    \end{center}
\end{minipage}
\\
\\
\end{tabular}
}
\end{minipage}
\end{figure*}


\subsection{Results and Discussion}
The following discusses the results obtained by running OPE simulations with 50 different logged datasets generated under different random seeds. We compare the estimators' accuracy by their MSE normalized by the true policy value of the target policy, which is defined as $\mse / (V(\pi))^2$. We use $L = 8$, $|\calA_l|=10\, (\forall l \in [L])$, and $n = 4000$ as default experiment parameters.

Figures~\ref{fig:slate_size_wiki} and~\ref{fig:slate_size_eurlex} report the estimators' MSE with varying slate sizes ($L$) and reward functions (1)-(3) on Wiki10-31K and Eurlex-4K, respectively. The results show that LIPS clearly outperforms the existing estimators across a range of slate sizes $(L \in \{4,6,\ldots,12\})$ and various non-linear reward functions. In contrast, we find that PI is likely to produce higher MSEs with growing slate sizes $L$ on some reward functions. This is due to the fact that it produces larger variance when the slate size becomes larger. It is also true that the violation of linearity is likely to produce larger bias with larger slate sizes where the interaction effects between slots could be larger. DM often performs worse than LIPS due to its high bias that arises from the estimation error of its reward predictor $\hat{q}$. We also observe that LIPS substantially outperforms IPS and MIPS due to substantially reduced variance via latent importance weighting. The observation that LIPS consistently performs much better than MIPS is particularly intriguing since MIPS uses a sufficient abstraction based on the knowledge about the true reward functions. This observation empirically demonstrates that LIPS and its associated optimization procedure can strategically exploit an \textit{insufficient} slate abstraction, leading to substantial variance reduction while introducing only a small amount of bias. Moreover, the comparison between LIPS and LIPS (w/ best $\beta$) implies that the feasible procedure of tuning the hyper-parameter $\beta$ via SLOPE~\citep{su2020adaptive} is often near-optimal even though it uses only observable data $\calD$, while LIPS can still be slightly improved with a more refined tuning method.

We also obtain similar observations in Figures~\ref{fig:data_size_wiki} and ~\ref{fig:data_size_eurlex} where we compare the estimators with varying data sizes from $n = 1000$ to $n=16000$ on the Wiki10-31K and Eurlex-4K datasets. We observe that PI struggles particularly when the sample size is small due to its high variance and the linearity assumption. DM is highly biased on both datasets and performs even worse than PI in many cases because estimating the reward function in large slate spaces is particularly challenging. IPS and MIPS produce much larger variance than other methods due to slate-wise importance weighting, and we see that they do not converge even with the largest data size, suggesting that they need even larger logged dataset to be effective. Finally, LIPS performs much better than other estimators in most cases with a feasible hyper-parameter selection by SLOPE, providing a further argument about its effectiveness for non-linear reward functions.

Beyond the basic comparisons against DM, IPS, PI, and MIPS, we finally compare LIPS with a set of hybrid estimators including DR~\citep{dudik2011doubly}, PI-DR~\citep{su2020doubly}, and OffCEM~\citep{saito2023off} (Appendix~\ref{app:experiment} defines these estimators in detail with math notations). Figures~\ref{fig:slate_size_wiki_dr} and~\ref{fig:slate_size_eurlex_dr} show this comparison on Wiki10-31K and Eurlex-4K for varying slate sizes $(L \in \{4,6,\ldots,12\})$. Figures~\ref{fig:data_size_wiki_dr} and~\ref{fig:data_size_eurlex_dr} compare the methods on the datasets for varying data sizes ($n \in \{1000,2000,\ldots,16000\}$). The results demonstrate that LIPS outperforms these hybrid estimators for a range of experiment values ($L$ and $n$) and non-linear reward functions. This is because DR has the variance issue due to its slate-wise importance weighting while PI-DR suffers from bias due to its linearity assumption and variance due to its slot-wise importance weighting. OffCEM is an extension of MIPS and uses a sufficient abstraction, but it is not optimized towards the MSE, and thus produces much larger variance than LIPS. These empirical observations suggest that, for OPE of slate bandits, tuning the definition of importance weights by optimizing slate abstractions with our method 
is more crucial and effective than introducing a reward estimator ($\hat{q}$) as in the hybrid estimators.

\section{Conclusion and Future Work}
This paper studied OPE of slate contextual bandits where existing estimators encounter significant challenges in terms of bias and variance due to combinatorial action spaces and restrictive assumptions on the reward function. To overcome these limitations, we proposed LIPS, a novel estimator that is built on slate abstraction to substantially reduce variance. Our analysis demonstrates that LIPS can be unbiased given a sufficient abstraction, and provides a substantial reduction in variance. We also observed that the advantages of LIPS may be maximized when we intentionally use an insufficient abstraction. Based on this analysis, we presented a method for optimizing slate abstraction to minimize the MSE of LIPS, which differentiates our work from MIPS~\citep{saito2022off} and OffCEM~\citep{saito2023off}.

As future work, it would be practically valuable to incorporate control variates into LIPS to further improve its variance, which is considered a non-trivial extension since it would change the way we should optimize slate abstractions. It would also be valuable to perform real-world experiments with slate bandit data from live enviornments once such a public dataset becomes available.

\clearpage
\begin{acks}
We would like to thank Tatsuhiro Shimizu for their helpful feedback. Additionally, we would like to thank the anonymous reviewers for their constructive reviews and discussions.
\end{acks}

\bibliographystyle{ACM-Reference-Format}
\balance
\bibliography{ref.bib}

\appendix
\clearpage
\onecolumn

\begin{figure*}[tb]
\centering
\begin{minipage}[b]{0.95\linewidth}
\begin{algorithm}[H]
\caption{Optimization procedure for slate abstraction used in LIPS} \label{algo:abstraction} 
  \begin{algorithmic}[1]
    \REQUIRE logged data $\calD$, bias-variance tradeoff hyper-parameter $\beta$, learning rates for slate abstraction model $\tau_{\theta}$, slate reconstruction model $\tau_{\psi}$, and reward construction model $\tau_{\omega}$, maximum gradient steps $T$, batch size $B$
    \ENSURE optimized slate abstraction distribution $p_{\theta}(z \,|\, \mx, \ms)$
    \STATE Initialize the parameters of the slate abstraction model, slate reconstruction model, and reward construction model $(\theta, \psi, \omega)$
    \FOR{$t \in \{1, 2, \cdots, T\}$}
        \STATE Sample size $B$ of mini-batch data $\calD_B^{(t)} \sim \calD$
        \FOR{$i \in \{1, 2, \cdots, B\}$}
            \STATE Retrieve data tuple $(\mx_i, \ms_i, r_i) \sim \calD_B^{(t)}$
            \STATE Sample a slate abstraction as $z_i \sim p_{\theta}(\cdot \,|\, \mx_i, \ms_i)$
        \ENDFOR
        \STATE Compute the slate reconstruction loss as \\
        \quad \quad $\hat{\mathcal{L}}_{b1}(\mx, \ms, \pi_0; \theta,\psi) = \frac{1}{B} \sum_{i \in [B]} \mathrm{log} \, p_{\psi}(\ms_i \,|\, \mx_i, z_i)$
        \STATE Compute the reward construction loss as \\
        \quad \quad $\hat{\mathcal{L}}_{b2}(\mx, \ms, \pi_0; \theta,\omega) = \frac{1}{B} \sum_{i \in [B]} (r_i - \hat{q}_{\omega}(\mx_i, z_i))^2$
        \STATE Compute the KL (regularization) loss as \\
        \quad \quad $\hat{\mathcal{L}}_{v1}(\mx, \ms, \pi_0; \theta) = \frac{1}{B} \sum_{i \in [B]} (\mathrm{log} \, p_{\theta}(z_i | \mx_i, \ms_i) - C)$ \\
        \quad \quad (where $C = - \mathrm{log}(|\calZ|)$ is the log of the prior distribution of $p_{\psi}(z | \mx, \ms) = |\calZ|^{-1}$)
        \STATE Compute the total loss as \\
        \quad \quad $\hat{\mathcal{L}}(\mx, \ms, \pi_0; \theta, \psi, \omega) = \hat{\mathcal{L}}_{b1}(\mx, \ms, \pi_0; \theta,\psi) + \hat{\mathcal{L}}_{b2}(\mx, \ms, \pi_0; \theta,\omega) - \beta \hat{\mathcal{L}}_{v1}(\mx, \ms, \pi_0; \theta)$
        \STATE Update the parameters of the slate abstraction and reconstruction models \\
        \quad \quad $\theta_{t} \leftarrow \theta_{t - 1} + \tau_{\theta} \, \nabla_{\theta} \hat{\mathcal{L}}(\mx, \ms, \pi_0; \theta,\psi,\omega)$ \\
        \quad \quad $\psi_{t} \leftarrow \psi_{t - 1} + \tau_{\psi} \, \nabla_{\psi} \hat{\mathcal{L}}(\mx, \ms, \pi_0; \theta,\psi,\omega)$
        \STATE Update the parameter of the reward construction model \\ 
        \quad \quad $\omega_{t} \leftarrow \omega_{t - 1} - \tau_{\omega} \, \nabla_{\omega} \hat{\mathcal{L}}_{b2}(\mx, \ms, \pi_0; \theta,\omega)$
    \ENDFOR
  \end{algorithmic}
\end{algorithm}
\end{minipage}
\vspace{8mm}
\end{figure*}

\section{Experimental details and additional results} \label{app:experiment}
We describe the detailed experimental settings and additional results omitted in the main text.

\subsection{The Baseline Estimators}
Below, we summarize the definition and properties of the baseline estimators. \\

\noindent \textbf{Direct Method (DM)~\citep{beygelzimer2009offset}} \; 
DM is a model-based approach, which first obtains a reward predictor $\hat{q}(\mx, \ms) \approx \mE[r|\mx,\ms]$ and then estimates the policy value of $\pi$ using $\hat{q}$ as follows.
\begin{align*}
    \hat{V}_{\text{DM}}(\pi; \calD) 
    := \frac{1}{n} \sum_{i=1}^n \sum_{\ms \in \calS} \pi(\ms | \mx_i) \hat{q}(\mx_i, \ms) \; 
    \left(= \frac{1}{n} \sum_{i=1}^n \mathbb{E}_{\ms \sim \pi(\ms | \mx_i)} [ \hat{q}(\mx_i, \ms) ] \right)
\end{align*}
The accuracy of DM depends on the accuracy of $\hat{q}(\mx, \ms)$, and the prediction error of $\hat{q}(\mx, \ms)$ introduces bias and makes DM no longer consistent.
Oftentimes, DM has high bias because $\hat{q}(\mx, \ms)$ needs to be trained on logged data with partial feedback. \\

\noindent \textbf{Inverse Propensity Scoring (IPS)~\citep{strehl2010learning}} \; As described in the main text, IPS applies the importance sampling technique to reweigh the observed rewards and correct the distribution shift between $\pi$ and $\pi_0$ as follows.
\begin{align*}
    \hat{V}_{\text{IPS}}(\pi; \calD) 
    := \frac{1}{n} \sum_{i=1}^n \frac{\pi(\ms_i | \mx_i)}{\pi_0(\ms_i | \mx_i)} r_i 
    = \frac{1}{n} \sum_{i=1}^n \left( \prod_{l=1}^L \frac{\pi(a_{i,l} | \mx_i)}{\pi_0(a_{i, l} | \mx_i)} \right) r_i
\end{align*}
IPS is unbiased under some identification assumptions. However, it has exponential variance due to the large slate space~\citep{swaminathan2017off}. \\

\noindent \textbf{PseudoInverse (PI)~\citep{swaminathan2017off}} \; PI assumes that the expected reward can be linearly attributed to each slot as $q(\mx, \ms) = \sum_{l=1}^L \phi_l(\mx, a_l)$, where $\{ \phi_l \}_{l=1}^L$ is some (latent) intrinsic reward function. Based on this assumption, PI corrects the distribution shift between $\pi$ and $\pi_0$ by applying slot-level importance sampling, as shown in the main text. 
\begin{align*}
    \hat{V}_{\text{PI}}(\pi; \calD) 
    := \frac{1}{n} \sum_{i=1}^n \left( \sum_{l=1}^L \frac{\pi(a_{i,l} | \mx_i)}{\pi_0(a_{i,l} | \mx_i)} - L + 1 \right) r_i
\end{align*}
PI is unbiased when the linearity assumption holds, while it can produce non-negligible bias when the assumption does not hold. In addition, while the variance of PI is much smaller than IPS, PI may still suffer from a high variance when there are many unique sub-actions and $\sum_{l=1}^L |\calA_l|$ becomes large~\citep{saito2022off}. \\

\noindent \textbf{Marginalized IPS (MIPS)~\citep{saito2022off}} \; MIPS is originally defined for a contextual bandit setting where action embeddings (like the genre of a video or sentiment of a movie) are observable. It operates under the following data generation process:
\begin{align}
    (\mx, a, \me, r) \sim p(\mx)\pi(a|\mx)p(\me|a)p(r|\mx,a,\me) \label{eq:embedding_generation}
\end{align}
In this scenario, $\mx \in \calX$ denotes the context, $a \in \calA$ represents an action, $\me \in \calE$ is an embedding vector, and $r \in \mathbb{R}$ signifies a reward. It is important to distinguish that these embeddings in Eq.~\eqref{eq:embedding_generation} differ from the abstractions in our main text formulation. First, embeddings are observable in logged data, while our abstractions are not. Moreover, embeddings come from some distribution $p(\me | \mx, \ma)$, unlike our problem setting where $p(z | \mx, \ms)$ is optimized.
Given this data generation process (Eq.~\eqref{eq:embedding_generation}), MIPS implements importance sampling within the embedding space $\calE$ as:
\begin{align*}
    \hat{V}_{\text{MIPS}}^{\text{(original)}} 
    := \frac{1}{n} \sum_{i=1}^n \frac{\pi(\me_i|\mx_i)} {\pi_0(\me_i|\mx_i)} r_i 
    = \frac{1}{n} \sum_{i=1}^n \frac{\sum_{a \in \calA} \pi(a | \mx_i) p(\me_i | a)} {\sum_{a \in \calA} \pi_0(a | \mx_i) p(\me_i | a)} r_i 
\end{align*}
where $\pi(\me | \mx) = \sum_{a \in \calA} \pi(a | \mx) p(\me | a)$ is the marginal probability of a policy $\pi$ chooses the actions associated with the embedding $\me$. MIPS is unbiased when the following \textit{no direct effect} assumption holds.
\begin{assumption} \textit{(Assumption 3.2 of \citet{saito2022off})} Action $a$ is said to have no direct effect on the reward if $a \perp r \,|\, x, \me$.
\end{assumption} \label{assm:no_direct_effect}

\noindent Under Assumption~\ref{assm:no_direct_effect}, the reward distribution can be expressed as $p(r|x,a,e) = p(r|x,e)$. While (slate) embeddings are unavailable in our setting, a \textit{sufficient slate abstraction} defined in Definition~\ref{def:sufficient} satisfies the no direct effect assumption (Assumption~\ref{assm:no_direct_effect}) if we regard $\phi(\ms)$ as an embedding. Therefore, in our experiment, we regard the following estimator, which satisfies the no direct effect assumption, as MIPS, given that $q(\mx, \ms)$ depends only on $\tilde{\ms} = (a_1, a_2, \cdots, a_{\lfloor L / 2 \rfloor})$.
\begin{align*}
    \hat{V}_{\text{MIPS}}(\pi; \calD) 
    := \frac{1}{n} \sum_{i=1}^n \frac{\pi(\tilde{\ms}_i | \mx_i)}{\pi_0(\tilde{\ms}_i | \mx_i)} r_i \; \left 
    (= \frac{1}{n} \sum_{i=1}^n \left( \prod_{l=1}^{\lfloor L/2 \rfloor} \frac{\pi(a_{i,l} | \mx_i)}{\pi_0(a_{i, l} | \mx_i)} \right) r_i \right)
\end{align*}
This version of MIPS is unbiased and reduces the variance compared to IPS. However, its variance remains high because the sufficient slate space of $\tilde{\ms}$ is still extremely large. \\

\noindent \textbf{Doubly Robust (DR)~\citep{dudik2014doubly}} \; 
DR is a hybrid estimator that combines both model-based and importance sampling-based approaches. Specifically, DR uses $\hat{q}(\mx, \ms)$ as a control variate and applies importance sampling only to the residual of $\hat{q}$ to reduce the variance as follows.
\begin{align*}
    \hat{V}_{\text{DR}}(\pi; \calD) 
    &:= \frac{1}{n} \sum_{i=1}^n \left\{ \frac{\pi(\ms_i | \mx_i)}{\pi_0(\ms_i | \mx_i)} (r_i - \hat{q}(\mx_i, \ms_i)) + \sum_{\ms \in \calS} \pi(\ms | \mx_i) \hat{q}(\mx_i, \ms) \right\} \\ 
    &= \frac{1}{n} \sum_{i=1}^n \left( \prod_{l=1}^L \frac{\pi(a_{i,l} | \mx_i)}{\pi_0(a_{i, l} | \mx_i)} \right) (r_i - \hat{q}(\mx_i, \ms_i)) + \hat{V}_{\text{DM}}(\pi; \calD)
\end{align*}
DR is unbiased and reduces the variance of IPS when $|\hat{q}(\mx, \ms) - q(\mx, \ms)| \leq q(\mx, \ms)$ holds for any $(\mx, \ms) \in \calX \times \calS$. However, DR can still suffer from high variance when the importance weight is large or $\hat{q}(\mx, \ms)$ is inaccurate~\citep{saito2022off}. \\

\noindent \textbf{PI-DR~\citep{su2020doubly,vlassis2021control}} \; 
PI-DR is a DR-variant of PI, which estimates the policy value as follows.
\begin{align*}
    \hat{V}_{\text{PI-DR}}(\pi; \calD) 
    &:= \frac{1}{n} \sum_{i=1}^n \left\{ \left( \sum_{l=1}^L \frac{\pi(a_{i,l} | \mx_i)}{\pi_0(a_{i,l} | \mx_i)} - L + 1 \right) (r_i - \hat{q}(\mx_i, \ms_i)) + \sum_{\ms \in \calS} \pi(\ms | \mx_i) \hat{q}(\mx_i, \ms) \right\} \\
    &= \frac{1}{n} \sum_{i=1}^n \left( \sum_{l=1}^L \frac{\pi(a_{i,l} | \mx_i)}{\pi_0(a_{i,l} | \mx_i)} - L + 1 \right) (r_i - \hat{q}(\mx_i, \ms_i)) + \hat{V}_{\text{DM}}(\pi; \calD)
\end{align*}
PI-DR is unbiased when the linearity assumption holds. However, its bias becomes high when the assumption does not hold. PI-DR also reduces the variance of PI under a reasonable assumption about the accuracy of reward prediction ($|\hat{q}(\mx, \ms) - q(\mx, \ms)| \leq q(\mx, \ms)$), while the variance problem can remain when $\sum_{l=1}^L |\calA_l|$ is large or $\hat{q}(\mx, \ms)$ is inaccurate.

\noindent \textbf{OffCEM~\citep{saito2023off}} \; 
OffCEM is another hybrid estimator that combines model-based and importance sampling-based approaches building on the MIPS estimator as follows.
\begin{align*}
    \hat{V}_{\text{OffCEM}}(\pi; \calD) 
    &:= \frac{1}{n} \sum_{i=1}^n \left\{ \frac{\pi(\tilde{\ms}_i | \mx_i)}{\pi_0(\tilde{\ms}_i | \mx_i)} (r_i - \hat{q}(\mx_i, \ms_i)) + \sum_{\ms \in \calS} \pi(\ms | \mx_i) \hat{q}(\mx_i, \ms) \right\} \\
    &= \frac{1}{n} \sum_{i=1}^n \left( \prod_{l=1}^{\lfloor L/2 \rfloor} \frac{\pi(a_{i,l} | \mx_i)}{\pi_0(a_{i, l} | \mx_i)} \right) (r_i - \hat{q}(\mx_i, \ms_i))  + \hat{V}_{\text{DM}}(\pi; \calD)
\end{align*}
OffCEM is unbiased either when \textit{no direct effect} assumption holds about $\tilde{\ms}$ or when the reward predictor accurately captures the pair-wise value difference between two slates within the same slate cluster (i.e., $\hat{q}(\mx, \ms_1) - \hat{q}(\mx, \ms_2) = q(\mx, \ms_1) - q(\mx, \ms_2), \forall \ms_1, \ms_2, \tilde{\ms}_1 = \tilde{\ms}_2$). OffCEM also reduces the variance of DR, however, the degree of variance reduction remains small when the slate space of $\tilde{s}$ remains large.

\subsection{Hyperparameter tuning via the SLOPE algorithm~\citep{su2020adaptive, tucker2021improved}} \label{app:slope}
To tune the bias-variance tradeoff hyperparameter $\beta$ of our optimization procedure based only on the logged bandit data $\calD$, we use the SLOPE algorithm~\citep{su2020adaptive,tucker2021improved}. SLOPE is able to identify a suitable hyperparameter $\lambda$ for an OPE estimator from a candidate set $\Lambda := \{ \lambda_m \}_{m=1}^M$ as long as an estimator satisfies the following \textit{monotonicity} condition~\citep{tucker2021improved}. 
\begin{enumerate}
    \item $\mathrm{Bias}(\hat{V}(\cdot; \, \lambda_m)) \leq \mathrm{Bias}(\hat{V}(\cdot; \, \lambda_{m+1})), \forall m \in [M - 1]$
    \item $\mathrm{CNF}(\hat{V}(\cdot; \, \lambda_m)) \geq \mathrm{CNF}(\hat{V}(\cdot; \, \lambda_{m+1})), \forall m \in [M - 1]$
\end{enumerate}
where $\mathrm{CNF}(\cdot)$ is a high probability bound on the deviation of $\hat{V}$ such as the Hoeffding and Bernstein bounds~\citep{thomas2015confidence}. Note that, LIPS and its hyper-parameter $\beta$ satisfy this monotonicity condition, as we know that a larger value of $\beta$ reduces the variance more, while a smaller value increasingly reduces the bias. Specifically, SLOPE selects the hyperparameter as follows.
\begin{align*}
    \hat{m} 
    &:= \max \{m \in [M] : |\hat{V}(\cdot; \, \lambda_m) - \hat{V}(\cdot; \, \lambda_{m'})| \\
    & \quad \quad \quad \quad \quad \quad \quad \quad \quad \quad \leq \mathrm{CNF}(\hat{V}(\cdot; \lambda_m)) + (\sqrt{6} - 1) \mathrm{CNF}(\hat{V}(\cdot; \, \lambda_{m'})), \forall m' < m \}.
\end{align*}
When the monotonicity condition holds, SLOPE guarantees that the deviation of $\hat{V}$ is upper bounded as below with probability $1 - \delta$:
\begin{align*}
    |\hat{V}(\cdot; \, \lambda_{\hat{m}}) - \hat{V}(\cdot; \, \lambda_{m^{\ast}})| \leq (\sqrt{6} + 3) \min_{m \in [M]} (\mathrm{Bias}(\hat{V}(\cdot; \, \lambda_m)) + \mathrm{CNF}(\hat{V}(\cdot; \, \lambda_m))),
\end{align*}
where $\hat{m}$ is the selected hyperparameter and $m^{\ast}$ is the best hyperparameter among the candidate set. Even when the condition does not hold true, SLOPE guarantees the following looser bound.
\begin{align*}
    |\hat{V}(\cdot; \, \lambda_{\hat{m}}) - \hat{V}(\cdot; \, \lambda_{m^{\ast}})| \leq (\sqrt{6} + 3) \min_{m \in [M]} (\max_{j \leq m} \mathrm{Bias}(\hat{V}(\cdot; \, \lambda_j)) + \max_{k \leq m} \mathrm{CNF}(\hat{V}(\cdot; \, \lambda_k))).
\end{align*}
We refer the reader to \citep{tucker2021improved} for the detailed theoretical analysis of SLOPE.

\begin{table}[t]
    \centering
    \caption{Statistics of the Extreme Classification datasets used in the experiments.}
    \vspace{-3mm}
    \label{tab:data-stats}
    \begin{tabular}{c||c|c|c|c}
        \toprule
        dataset & \# of documents & features of documents & \# of labels & avg. labels per document
         \\ \midrule \midrule
        Wiki10-31K & 14,146	(6,616) & raw texts & 30,938 & 18.64 \\ 
        Eurlex-4K & 15,539 (3,809) & raw texts & 3,993 & 5.31 \\
        Delicious & 12,920 (3,185) & 500 dim. of BoW & 983 & 19.03 \\
        \bottomrule
    \end{tabular}
    \vskip 0.1in
    \raggedright
    \fontsize{9pt}{9pt}\selectfont \textit{Note}:
    The column ``\# of documents" describes ``\# of train samples (\# of test samples)" of the datasets. We use the training dataset for performing slate OPE, while the test set is used for training the base classifier $\tilde{q}$ to form a logging policy. The raw texts of the Wiki10-31K and Eurlex-4K datasets are converted to 20-dimensional feature vectors via SentenceTransformer~\citep{reimers2019sentence} and PCA~\citep{abdi2010principal}. Finally, for Wiki10-31K and Eurlex-4K, we extract the top 1,000 dense labels after removing the labels that are relevant to more than 1,000 documents.
    \label{tab:runtime_slate_size}
    \vspace{5mm}
\end{table}

\begin{figure}
\centering
\begin{minipage}[c]{0.33\linewidth}
    \centering
    \includegraphics[clip, width=4.0cm]{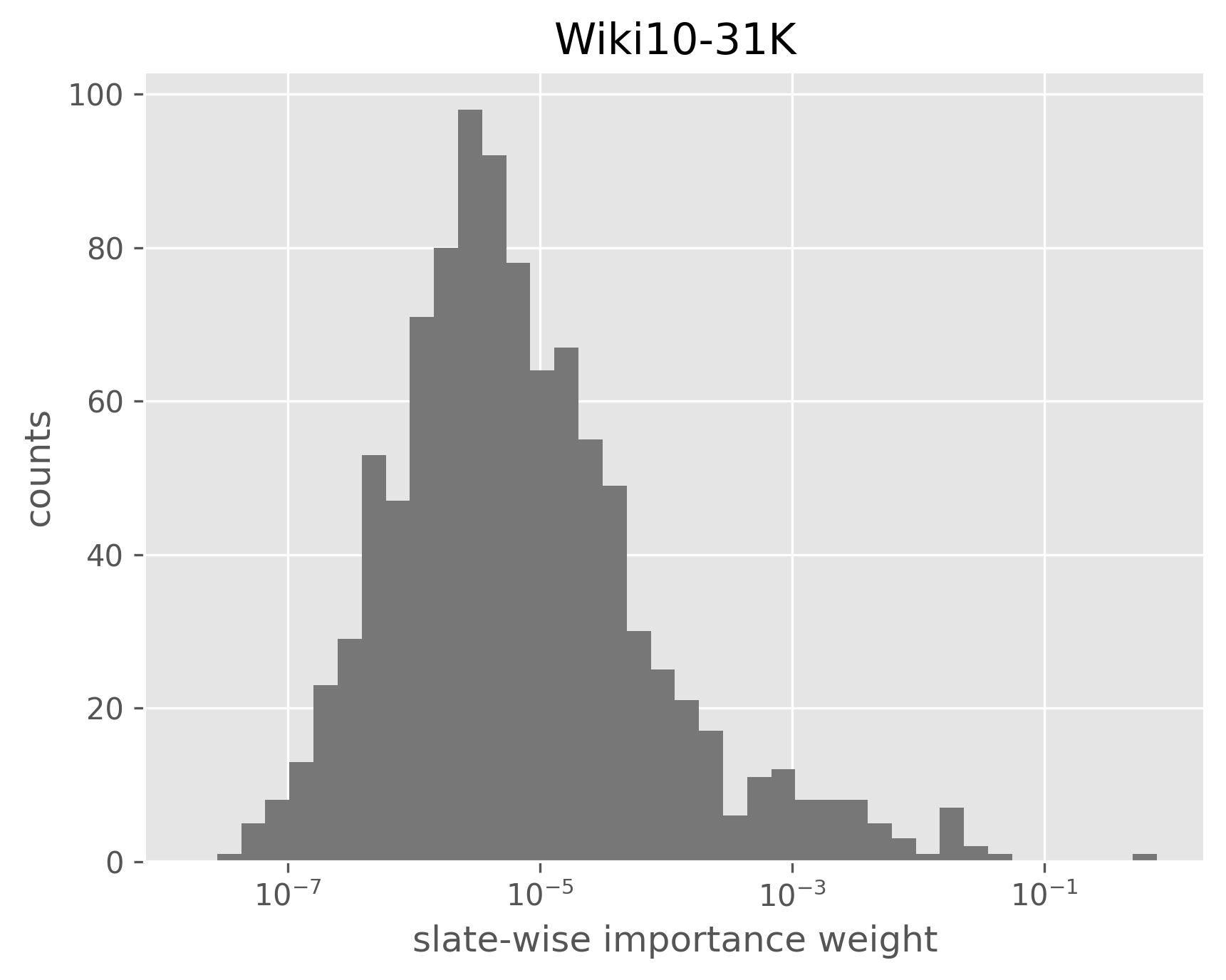}
    \vspace{1mm}
\end{minipage}
\begin{minipage}[c]{0.33\linewidth}
    \centering
    \includegraphics[clip, width=4.0cm]{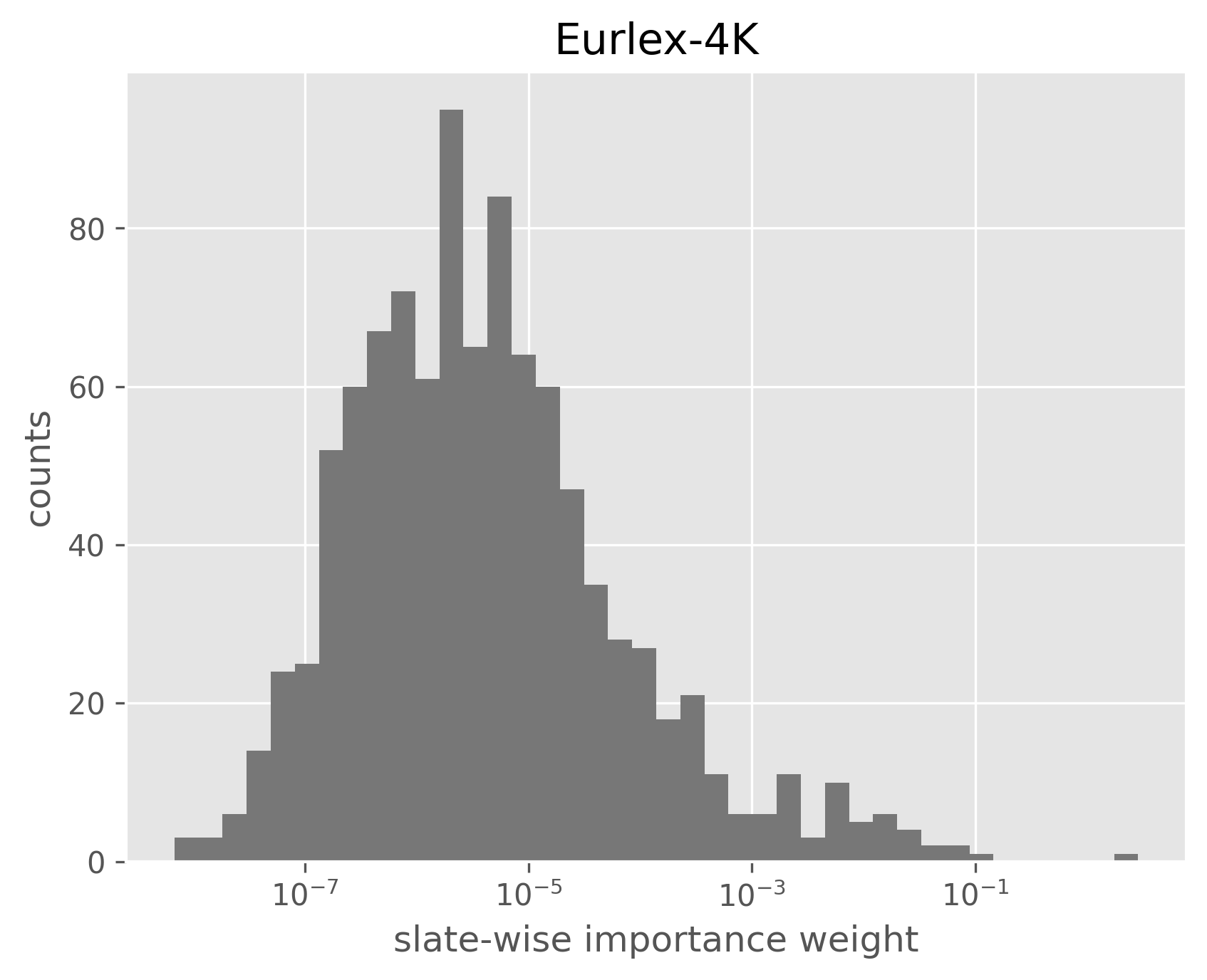}
    \vspace{1mm}
\end{minipage}
\begin{minipage}[c]{0.33\linewidth}
    \centering
    \includegraphics[clip, width=4.0cm]{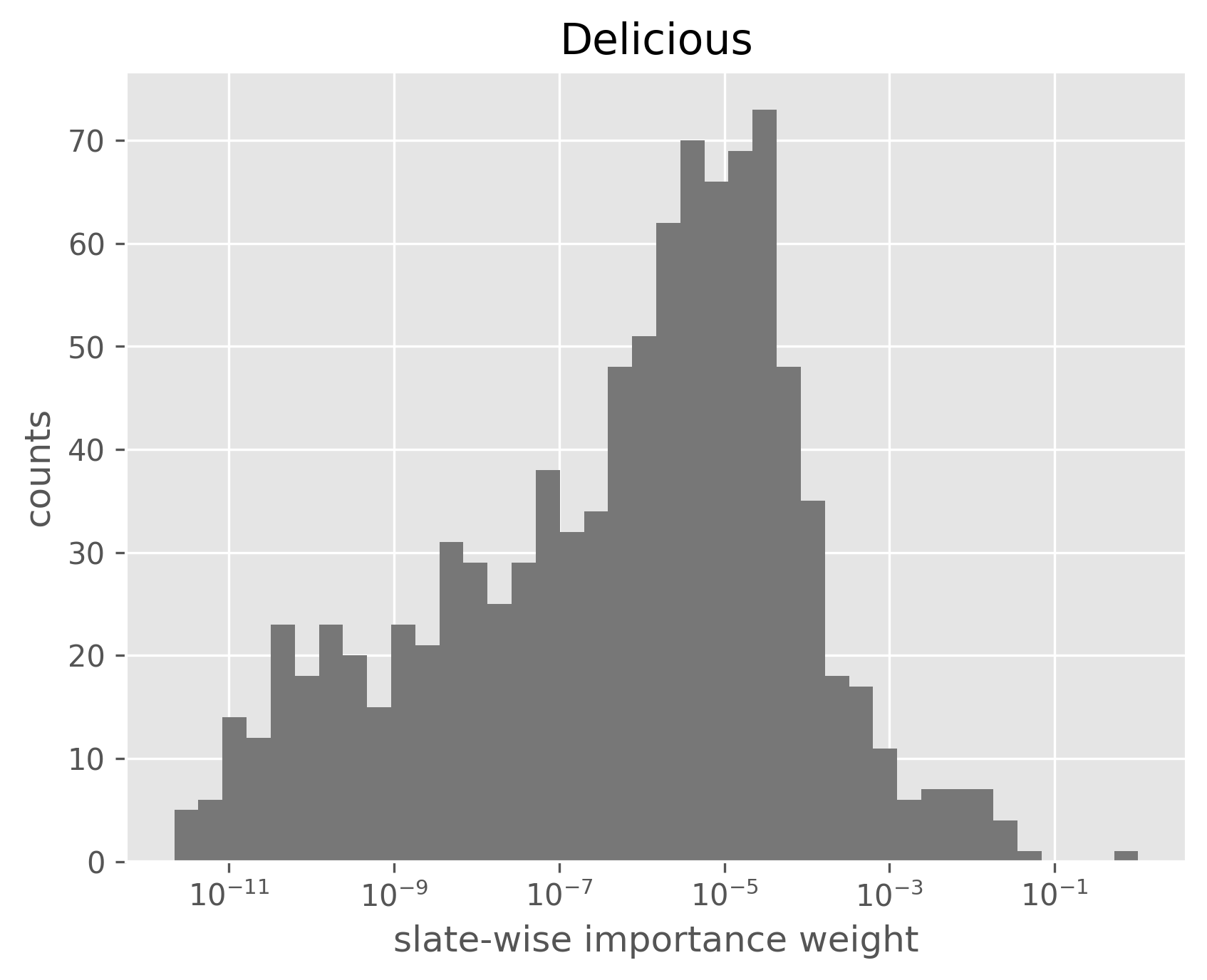}
    \vspace{1mm}
\end{minipage}
\caption{Empirical distributions of the slate-wise importance weight $w(\mx,\ms)$ for each dataset used in the experiment; (Left) Wiki10-31K (Center) Eurlex-4K (Right) Delicious.} \label{fig:pscore}
\vspace{3mm}
\end{figure}

\subsection{Models and parameters}
We use a neural network with a 100-dimensional hidden layer to obtain the reward predictor $\hat{q}$ for DM. This network employs the Adam optimizer~\citep{kingma2014adam} to minimize the MSE loss, using a learning rate of 1e-2. The logged dataset is split, with 80\% used for training and the remaining 20\% for testing. The model is trained on the training set for 500 epochs, taking 10 gradient steps per epoch. An early stopping strategy is applied to $\hat{q}$, triggered when the test loss increases for five consecutive epochs. Due to the high computational cost of exact expectation over the slate space ($\calS$), we approximate the expectation over $\pi(\ms | \mx)$ by sampling 1,000 slates according to $\pi(\ms | \mx)$ for each context $\mx$.

For optimizing slate abstraction in the LIPS estimator, we use neural networks with a 100-dimensional hidden layer to parameterize the reward construction, slate abstraction, and slate reconstruction models. These models are also optimized using Adam~\citep{kingma2014adam}, with a learning rate of 1e-5 for both the Wiki10-31K and Eurlex-4K datasets. In all experiments, the reward loss is re-scaled by multiplying it by 100 to align its scale with that of the slate reconstruction loss. Initially, the models are trained for 1000 epochs with $\beta = 0.01$, followed by a tuning phase of 500 epochs for different $\beta$ values (0.1, 1.0, and 10.0). These experiments are conducted on an M1 MacBook Pro, and the optimization process for LIPS takes approximately 3 minutes for a single value of $\beta$ when $L=4$ and $n=4000$. For $L=12$, the computation time increases to about 4.5 minutes, demonstrating a sub-linear increase in runtime.

\subsection{Additional experiments on the Delicious dataset}
In addition to the experiments described in the main text, we also conducted comparisons using the Delicious dataset from the Extreme Classification repository~\citep{bhatia16}. This dataset comprises documents each associated with about 1,000 labels. The detailed dataset statistics are presented in Table~\ref{tab:data-stats}. We employed the same "supervised-to-bandit" conversion procedure as in the main text, with the key difference being the use of 500-dimensional Bag of Words (BoW) features for contexts, due to the absence of raw texts in the Delicious dataset.

\begin{figure*}[t]
\begin{minipage}[c]{0.99\hsize}
\centering
\scalebox{0.95}{
\begin{tabular}{c}
\begin{minipage}{0.80\hsize}
\begin{center}
\includegraphics[width=0.95\linewidth]{figs/label.png}
\vspace{-2mm}
\end{center}
\end{minipage}
\\
\\
\begin{minipage}{0.99\hsize}
    \begin{center}
        \includegraphics[clip, width=0.95\linewidth]{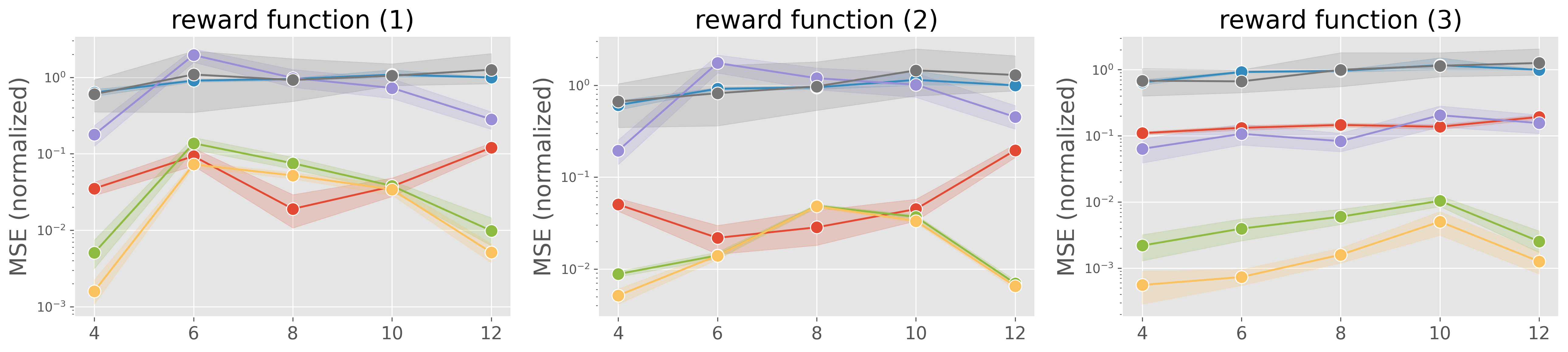}
        \vspace{-2mm}
        \caption{Comparison of the estimators' MSE (normalized by the true value $V(\pi)$) with varying slate sizes ($L$) and with reward functions (1) - (3) on the Delicious dataset. The y-axis is on a log scale.}
        \label{fig:slate_size_delicious}
        \vspace{3mm}
    \end{center}
\end{minipage}
\\
\\
\begin{minipage}{0.80\hsize}
\begin{center}
\includegraphics[width=0.95\linewidth]{figs/label_dr.png}
\vspace{-2mm}
\end{center}
\end{minipage}
\\
\\
\begin{minipage}{0.99\hsize}
    \begin{center}
        \includegraphics[clip, width=0.95\linewidth]{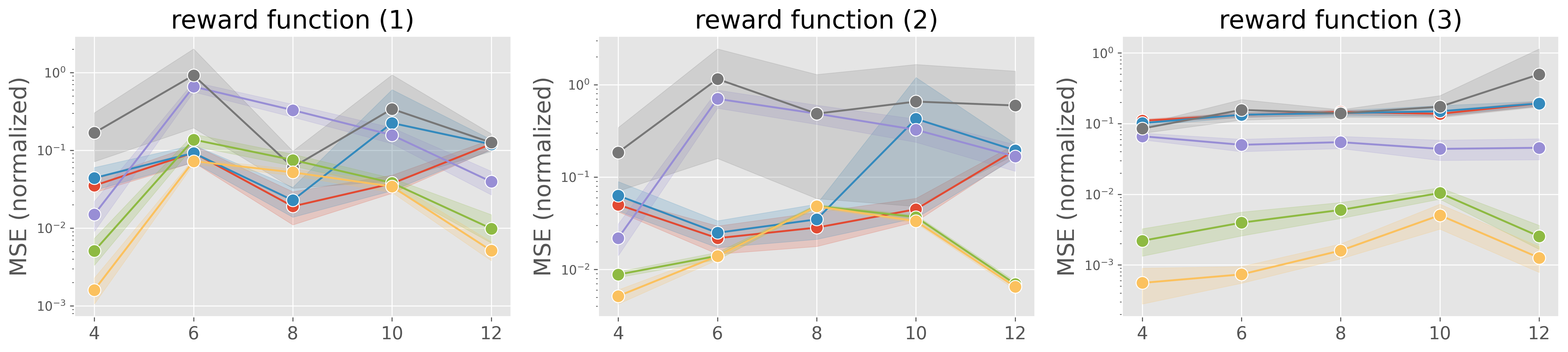}
        \vspace{-2mm}
        \caption{Comparing the LIPS' MSE (normalized by the true value $V(\pi)$) with those of DR estimators under varying slate sizes ($L$) and with reward functions (1) - (3) on the Delicious dataset. The y-axis is on a log scale.}
        \label{fig:slate_size_delicious_dr}
        \vspace{3mm}
    \end{center}
\end{minipage}
\\
\\
\begin{minipage}{0.80\hsize}
\begin{center}
\includegraphics[width=0.95\linewidth]{figs/label.png}
\vspace{-2mm}
\end{center}
\end{minipage}
\\
\\
\begin{minipage}{0.99\hsize}
    \begin{center}
        \includegraphics[clip, width=0.95\linewidth]{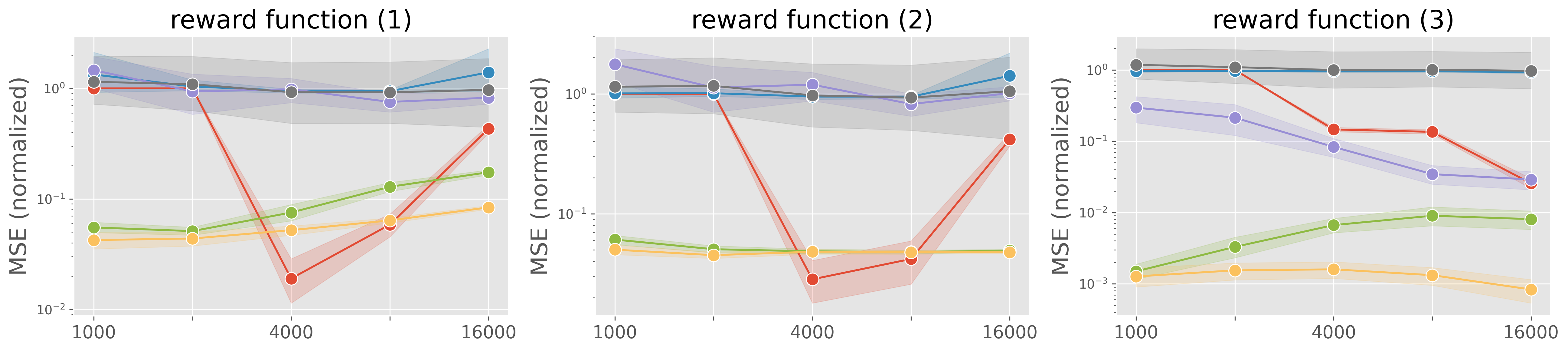}
        \vspace{-2mm}
        \caption{Comparison of the estimators' MSE (normalized by the true value $V(\pi)$) with varying data sizes ($n$) and with reward functions (1) - (3) on the Delicious dataset. The y-axis is on a log scale.}
        \label{fig:data_size_delicious}
        \vspace{3mm}
    \end{center}
\end{minipage}
\\
\\
\begin{minipage}{0.80\hsize}
\begin{center}
\includegraphics[width=0.95\linewidth]{figs/label_dr.png}
\vspace{-2mm}
\end{center}
\end{minipage}
\\
\\
\begin{minipage}{0.99\hsize}
    \begin{center}
        \includegraphics[clip, width=0.95\linewidth]{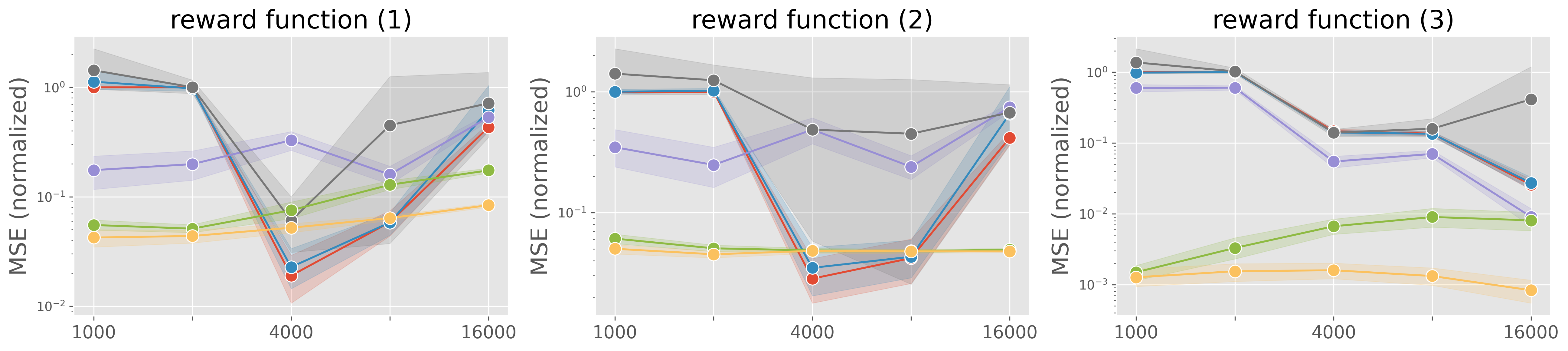}
        \vspace{-2mm}
        \caption{Comparing the LIPS' MSE (normalized by the true value $V(\pi)$) with those of DR estimators under varying data sizes ($n$) and with reward functions (1) - (3) on the Delicious dataset. The y-axis is on a log scale.}
        \label{fig:data_size_delicious_dr}
        \vspace{3mm}
    \end{center}
\end{minipage}
\\
\\
\end{tabular}
}
\end{minipage}
\end{figure*}

The comparison results between LIPS and the IPS-based estimators (as shown in Figures~\ref{fig:slate_size_delicious} and \ref{fig:data_size_delicious}), as well as the DR-based estimators (Figures~\ref{fig:slate_size_delicious_dr} and \ref{fig:data_size_delicious_dr}), are displayed with variations in slate sizes $L$ (Figures~\ref{fig:slate_size_delicious} and \ref{fig:slate_size_delicious_dr}) and data sizes $n$ (Figures~\ref{fig:data_size_delicious} and \ref{fig:data_size_delicious_dr}) on the Delicious dataset. These results are similar to the trends observed in the main text, showing that LIPS (ours, with a data-driven choice of $\beta$) typically outperforms the other estimators. IPS, PI, and MIPS often yield high estimation errors, while DM's accuracy is contingent on chance accuracy of the reward prediction. These findings indicate LIPS's capability to adeptly manage the bias-variance tradeoff, thus minimizing the MSE across diverse datasets.

\subsection{Additional ablation experiments with varying values of $\beta$}
\begin{figure*}[t]
\begin{minipage}[c]{0.99\hsize}
\centering
\scalebox{0.95}{
\begin{tabular}{c}
\begin{minipage}{0.50\hsize}
\begin{center}
\includegraphics[width=1.025\linewidth]{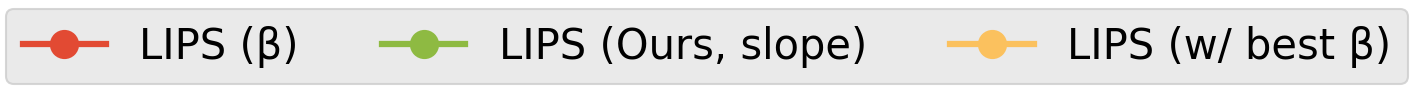}
\vspace{-5mm}
\end{center}
\end{minipage}
\\
\\
\begin{minipage}{0.99\hsize}
    \begin{center}
        \includegraphics[clip, width=1.02\linewidth]{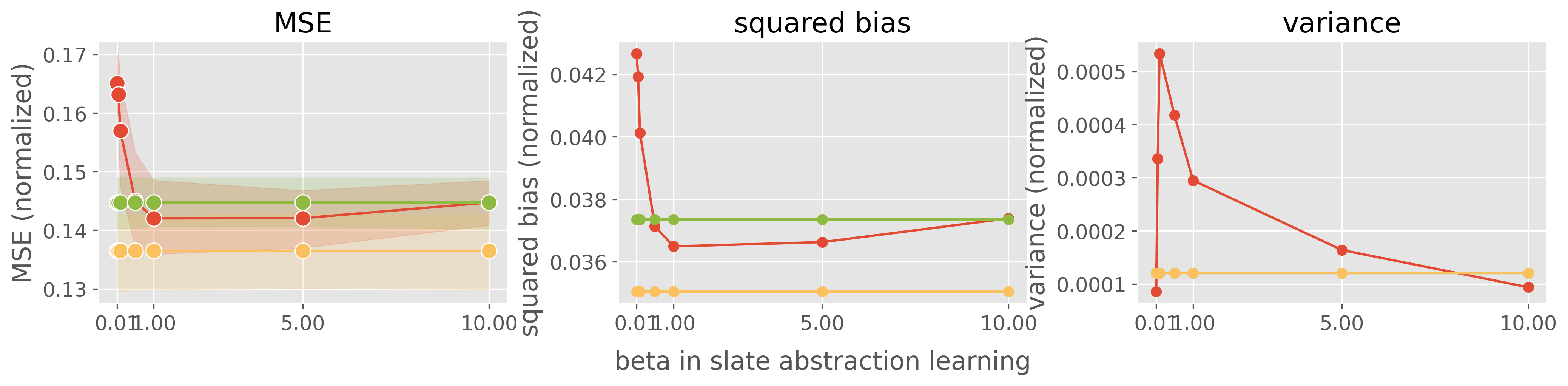}
        \vspace{-2mm}
        \caption{Comparison of the LIPS' MSE, squared bias, and variance (normalized by the true value $V(\pi)$) with varying values of $\beta$ with reward function (1) on the Wiki10-31K dataset}
        \label{fig:ablation_wiki}
        \vspace{3mm}
    \end{center}
\end{minipage}
\\
\\
\begin{minipage}{0.99\hsize}
    \begin{center}
        \includegraphics[clip, width=1.02\linewidth]{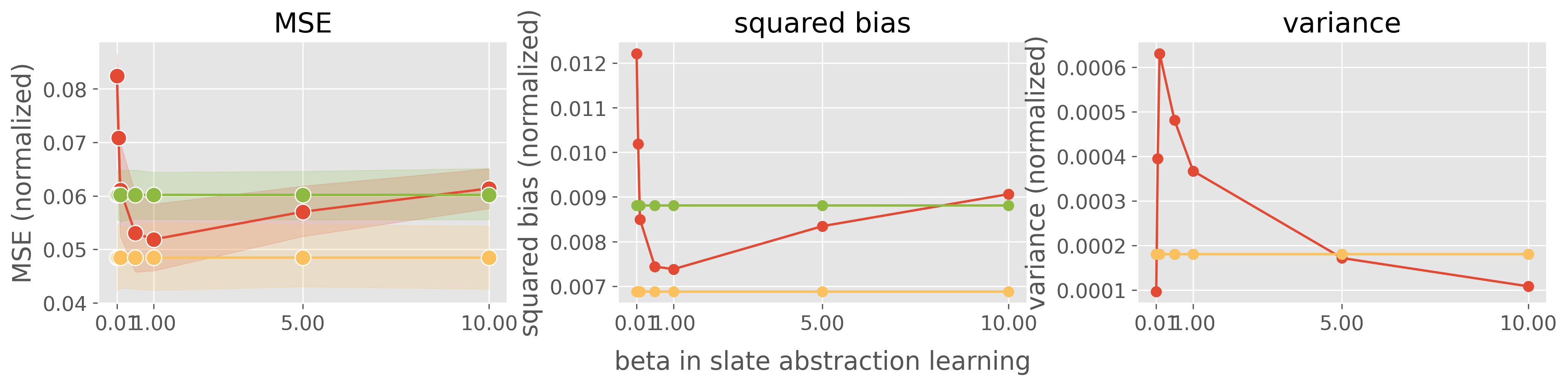}
        \vspace{-2mm}
        \caption{Comparison of the LIPS' MSE, squared bias, and variance (normalized by the true value $V(\pi)$) with varying values of $\beta$ with reward function (1) on the Eurlex-4K dataset.}
        \label{fig:ablation_eurlex}
    \end{center}
    \vspace{3mm}
\end{minipage}
\\
\\
\end{tabular}
}
\end{minipage}
\end{figure*}

We conducted an ablation study of LIPS with various $\beta$ values (\{0.01, 0.05, 0.1, 0.5, 1.0, 5.0, 10.0\}) on the Wiki10-31K and Eurlex-4K datasets, using a standard setting of $L=8$ and $n=4,000$. Figures \ref{fig:ablation_wiki} and \ref{fig:ablation_eurlex} illustrate the MSE, squared bias, and variance of LIPS for each $\beta$ value, as well as for the data-driven choice (LIPS (Ours)) and the best value of $\beta$, respectively. The results indicate a bias-variance tradeoff for $\beta$ values in the range \{1.0, 5.0, 10.0\}, and some instability for $\beta \leq 1.0$. Specifically, in both the Wiki10-31K and Eurlex-4K datasets, a smaller $\beta$ value (e.g., $\beta = 1.0$) leads to reduced bias but increased variance, while a larger $\beta$ value (e.g., $\beta = 5.0, 10.0$) further reduces variance but introduces more bias. These findings are consistent with our theoretical analysis, as discussed in Sections~\ref{sec:theoretical} and \ref{sec:abstraction}. Additionally, the results indicate potential further improvements regarding the data-driven selection of $\beta$. Nevertheless, our abstraction optimization enables accurate estimation across a range of $\beta$ values and demonstrates robust performance even with a data-driven $\beta$ selection, as detailed in the main text. This robustness is likely due to the ability of our slate abstraction model to identify an optimized abstraction that simultaneously reduces bias and variance.


\section{Detailed Bias-Variance Analysis of LIPS and Omitted Proofs} \label{app:proof}
Here, we provide the detailed discussion on the bias-variance properties of LIPS and the proofs of the theorems given in the main text.

\subsection{Preliminaries} \label{app:proof_prelim}

As a warm-up, we first derive an alternative expression of the latent importance weight as follows:
\begin{align}
    w_{\theta}(\mx, z)
    &= \frac{p_{\theta}(z \,|\, \mx,\pi)}{p_{\theta}(z \,|\, \mx,\pi_0)} \nonumber \\
    &= \frac{\sum_{\ms \in \calS} p_{\theta}(z\,|\,\mx,\ms) \pi(\ms\,|\,\mx)}{p_{\theta}(z \,|\, \mx, \pi_0)} \nonumber \\
    &= \frac{\sum_{\ms \in \calS} \frac{p_{\theta}(z\,|\,\mx,\pi_0) p_{\theta}(\ms\,|\,\mx,z,\pi_0)}{\pi_0(\ms\,|\,\mx)} \pi(\ms\,|\,\mx)}{p_{\theta}(z \,|\, \mx,\pi_0)} \nonumber \\
    &= \sum_{\ms \in \calS} p_{\theta}(\ms\,|\,\mx,z,\pi_0) \frac{\pi(\ms\,|\,\mx)}{\pi_0(\ms\,|\,\mx)} \nonumber \\
    &= \mE_{p_{\theta}(\ms\,|\,\mx,z,\pi_0)}[w(\mx,\ms)]
\end{align}
This means that $w_{\theta}(\mx, z)$ can be written as the conditional expectation of $w(\mx, \ms)$. Using this expression, we show how our optimization procedure for slate abstraction can balance the bias-variance tradeoff in the following subsection.

\subsection{Bias-variance control of LIPS with the hyper-parameter $\beta$}
In this section, we demonstrate that LIPS effectively interpolates between IPS and the naive average estimator (NAE). NAE is defined as the naive empirical average of the observed rewards: $\hat{V}_{\mathrm{NAE}}(\pi; \calD) := \frac{1}{n} \sum_{i = 1}^n r_i$. As NAE does not accommodate the distribution shift between $\pi_0$ and $\pi$, it often leads to a high bias. Contrarily, NAE achieves a much lower variance than IPS because it does not depend on importance weighting. As a result, IPS and NAE exhibit opposite characteristics in terms of the bias-variance tradeoff.

Next, given a sufficiently large abstraction space $\calZ$, LIPS becomes identical to IPS and NAE in the special cases as follows.
\begin{itemize}
    \item When $\beta = 0$, LIPS becomes equal to IPS.
    \item As $\beta \rightarrow +\infty$, LIPS gets close to NAE.
\end{itemize}
When $\beta = 0$, the abstraction optimization procedure ignores the regularization term (the third term). Therefore, LIPS optimizes a slate abstraction so that it can distinguish the slates accurately, leading to a one-to-one mapping between the slate and abstraction (i.e., both $p_{\theta}(z | \mx, \ms, \pi_0)$ and $p_{\psi}(\ms | \mx, z, \pi_0)$ becomes extremely close to either 0 or 1). This will make our latent importance weight equivalent to the slate-wise importance weight of IPS, and LIPS will be identical to IPS under such a condition. On the other hand, as $\beta \rightarrow +\infty$, the KL loss becomes more dominant. In such a situation, $p_{\theta}(z \,|\, \mx, \ms, \pi_0)$ becomes close to a (pre-defined) prior distribution $p_{\psi}(z \,|\, \mx, \pi_0)$, which leads to the following relationship about the slate abstraction distribution:
\begin{align*}
    p_{\theta}(\ms \,|\, \mx, z, \pi_0) = \pi_0(\ms | \mx) \frac{p_{\theta}(z \,|\, \mx, \ms, \pi_0)}{p_{\theta}(z \,|\, \mx, \pi_0)} \approx \pi_0(\ms | \mx) \frac{p_{\psi}^{(prior)}(z \,|\, \mx, \pi_0)}{p_{\psi}^{(prior)}(z \,|\, \mx, \pi_0)} = \pi_0(\ms | \mx)
\end{align*}
By combining this with the alternative expression of the latent importance weight derived in Appendix~\ref{app:proof_prelim}, we have the following.
\begin{align*}
    w_{\theta}(\mx, z) 
    = \mE_{p_{\theta}(\ms\,|\,\mx,z,\pi_0)}[w(\mx,\ms)]
    \approx \mE_{\pi_0(\ms \,|\, \ma)}[w(\mx,\ms)]
    = \sum_{\ms \in \calS} \pi_0(\ms \,|\, \mx) \frac{\pi(\ms \,|\, \mx)}{\pi_0(\ms \,|\, \mx)} = 1
\end{align*}
This means that LIPS applies no correction, i.e., $w_{\theta}(\mx, z) = 1$, when $\beta \rightarrow + \infty$. 

To summarize, by adjusting $\beta \in [0, +\infty)$, LIPS interpolates between IPS and NAE in a way that minimizes the MSE of LIPS, without introducing any structural assumptions on the reward function.

\subsection{Proof of Theorem~\ref{thrm:unbiased}} \label{app:unbiased}

The following provides the proof of Theorem~\ref{thrm:unbiased}.

\begin{proof} Given sufficient slate abstraction $\phi_{\theta}(\cdot)$ s.t., $\forall \ms \in \calS, q(\mx, \ms) = q(\mx, \phi_{\theta}(\ms))$, we have the following.
\begin{align*}
    \mE_{\calD}[\ours] 
    &= \mE_{\calD}[ w_{\theta}(\mx, \phi_{\theta}(\ms)) \, r ] \\
    &= \mE_{p(\mx)\pi_0(\ms\,|\,\mx)}[w_{\theta}(\mx, \phi_{\theta}(\ms)) \, q(\mx,\ms)] \\
    &= \mE_{p(\mx)\pi_0(\ms\,|\,\mx)}[w_{\theta}(\mx, \phi_{\theta}(\ms)) \, q(\mx,\phi_{\theta}(\ms))] \\
    &= \mE_{p(\mx)} \Bigl[ \sum_{\ms \in \calS} \pi_0(\ms | \mx)
    \frac{\pi(\phi_{\theta}(\ms) \,|\, \mx)}{\pi_0(\phi_{\theta}(\ms) \,|\, \mx)} q(\mx,\phi_{\theta}(\ms))\Bigr] \\
    &= \mE_{p(\mx)} \Bigl[ \sum_{\ms \in \calS} \pi_0(\phi_{\theta}(\ms) \,|\, \mx) \frac{\pi_0(\ms | \mx)}{\pi_0(\phi_{\theta}(\ms) \,|\, \mx)}
    \frac{\pi(\phi_{\theta}(\ms) \,|\, \mx)}{\pi_0(\phi_{\theta}(\ms) \,|\, \mx)} q(\mx,\phi_{\theta}(\ms))\Bigr] \\
    &= \mE_{p(\mx)} \Bigl[ \sum_{\ms \in \calS} \pi(\phi_{\theta}(\ms) \,|\, \mx) \pi_0(\ms \,|\, \mx, \phi_{\theta}(\ms)) \,
    q(\mx,\phi_{\theta}(\ms))\Bigr] \\
    &= \mE_{p(\mx)} \Bigl[ \sum_{z \in \calZ} \pi(z \,|\, \mx) \sum_{\ms \in \{\ms' \in\calS \mid \phi_{\theta}(\ms') = z\}} \pi_0(\ms \,|\, \mx, z) \,
    q(\mx,z)\Bigr] \\
    &= \mE_{p(\mx)} \Bigl[ \sum_{z \in \calZ} \pi(z \,|\, \mx)
    q(\mx,z)\Bigr] \\
    &= \mE_{p(\mx)} \Bigl[ \sum_{z \in \calZ} \pi(z \,|\, \mx) \sum_{\ms \in \{\ms' \in\calS \mid \phi_{\theta}(\ms') = z\}} \pi(\ms \,|\, \mx, z) \,
    q(\mx,z)\Bigr] \\
    &= \mE_{p(\mx)} \Bigl[ \sum_{\ms \in \calS} \pi(\phi_{\theta}(\ms) \,|\, \mx) \pi(\ms \,|\, \mx, \phi_{\theta}(\ms)) \,
    q(\mx,\phi_{\theta}(\ms))\Bigr] \\
    &= \mE_{p(\mx)} \Bigl[ \sum_{\ms \in \calS} \pi(\ms \,|\, \mx) \,
    q(\mx,\phi_{\theta}(\ms))\Bigr] \\
    &= \mE_{p(\mx)} \Bigl[ \sum_{\ms \in \calS} \pi(\ms \,|\, \mx) \,
    q(\mx,\ms)\Bigr] \\
    &= V(\pi)
    \end{align*}
\end{proof}
Note that we use $\phi_{\theta}(\ms) \, (= z) \in \calZ$ and $\calS = \{ \bigcup_{z \in \calZ} \bigcup_{\ms \in \{\ms' \in \calS \mid \phi_{\theta}(\ms') = z\}} \ms \}$.

\subsection{Proof of Theorem~\ref{thrm:bias}} \label{app:bias}

To prove Theorem~\ref{thrm:bias}, we first import the following lemma from \citep{saito2022off}.

\begin{lemma} \label{lemma:saito}
(Lemma B.1. of \citep{saito2022off}) For real-valued, bounded functions $f: \mathbb{N} \rightarrow \mathbb{R}, g: \mathbb{N} \rightarrow \mathbb{R}, h: \mathbb{N} \rightarrow \mathbb{R}$ where $\sum_{a \in [m]} g(a) = 1$, we have
\begin{align*}
    \sum_{a \in [m]} f(a) g(a) \Bigl( h(a) - \sum_{b \in [m]} g(b) h(b) \Bigr) = \sum_{a < b \leq m} g(a) g(b) (h(a) - h(b)) (f(a) - f(b))
\end{align*}
\end{lemma}

Then, we provide the proof in the following.

\begin{proof}
We show this in the case of a discrete slate abstraction. Similar proofs hold for a continuous slate abstraction by replacing $\sum_{z \in \calZ}$ for $\int_{z \in \calZ} dz$.

\begin{align*}
    & \mathrm{Bias}(\ours) \\
    &= \mE_{\calD}[w_{\theta}(\mx,z) \, r] - V(\pi) \\
    &= \mE_{p(\mx)\pi_0(\ms\,|\,\mx)p_{\theta}(z\,|\,\mx,\ms)}[w_{\theta}(x,z) q(\mx,\ms)] - \mE_{p(\mx)\pi(\ms\,|\,\mx)p_{\theta}(z\,|\,\mx,\ms)}[q(\mx,\ms)] \\
    &= \mE_{p(\mx)}\Bigl[\sum_{\ms \in \calS} \pi_0(\ms\,|\,\mx) \sum_{z \in \calZ} \frac{p_{\theta}(z\,|\,\mx,\pi_0)p_{\theta}(\ms\,|\,\mx,z,\pi_0)}{\pi_0(\ms\,|\,\mx)} w_{\theta}(\mx,z) q(\mx,\ms) \Bigl] \\
    & \quad - \mE_{p(\mx)}\Bigl[\sum_{\ms \in \calS} \pi(\ms\,|\,\mx) \sum_{z \in \calZ} \frac{p_{\theta}(z\,|\,\mx,\pi_0)p_{\theta}(\ms\,|\,\mx,z,\pi_0)}{\pi_0(\ms\,|\,\mx)} w_{\theta}(\mx,z) q(\mx,\ms) \Bigl] \\
    &= \mE_{p(\mx)}\Bigl[ \sum_{z \in \calZ} p_{\theta}(z\,|\,\mx,\pi_0) w_{\theta}(\mx, z) \sum_{\ms \in \calS} p_{\theta}(\ms\,|\,\mx,z,\pi_0) q(\mx,\ms) \Bigl] \\
    & \quad - \mE_{p(\mx)}\Bigl[ \sum_{z \in \calZ} p_{\theta}(z\,|\,\mx,\pi_0) \sum_{\ms \in \calS} p_{\theta}(\ms\,|\,\mx,z,\pi_0) w(\mx, \ms) q(\mx,\ms) \Bigl] \\
    &= \mE_{p(\mx)}\Bigl[ \sum_{z \in \calZ} p_{\theta}(z\,|\,\mx,\pi_0) \Bigl( \sum_{\ms \in \calS} p_{\theta}(\ms\,|\,\mx, z,\pi_0) w(\mx, \ms) \Bigr) \sum_{\ms' \in \calS} p_{\theta}(\ms'\,|\,\mx,z,\pi_0) q(\mx,\ms') \Bigl] \\
    & \quad - \mE_{p(\mx)}\Bigl[ \sum_{z \in \calZ} p_{\theta}(z\,|\,\mx,\pi_0) \sum_{\ms \in \calS} p_{\theta}(\ms\,|\,\mx,z,\pi_0) w(\mx, \ms) q(\mx,\ms) \Bigl] \\
    &= \mE_{p(\mx)p_{\theta}(z\,|\,\mx,\pi_0)}\Bigl[ \sum_{\ms \in \calS} p_{\theta}(\ms\,|\,\mx, z,\pi_0) w(\mx, \ms) \Bigl( \sum_{\ms' \in \calS} p_{\theta}(\ms'\,|\,\mx,z,\pi_0) q(\mx,\ms') - q(\mx, \ms) \Bigr) \Bigr] \\
    &= \mE_{p(\mx)p_{\theta}(z\,|\,\mx,\pi_0)}\Bigl[ \sum_{j < k \leq \,|\,\calS\,|\,} p_{\theta}(\ms_j\,|\,\mx,z,\pi_0) p_{\theta}(\ms_k\,|\,\mx,z,\pi_0) \\
    & \quad \quad \quad \quad \quad \quad \quad \quad \quad \quad \times (q(\mx,\ms_j) - q(\mx,\ms_k)) \times (w(\mx,\ms_k) - w(\mx,\ms_j)) \Bigr] 
\end{align*}
where the last line uses Lemma~\ref{lemma:saito}.
\end{proof}

\subsection{Proof of Theorem~\ref{thrm:variance}} \label{app:variance-reduction}
\begin{proof}
Here, we use the unbiasedness of LIPS under a sufficient slate abstraction (i.e., Theorem~\ref{thrm:unbiased}).
\begin{align*}
    & n \left( \mV_{\calD} [\ips] - \mV_{\calD} [\ours] \right) \\
    &= \mE_{\calD}[(w(\mx, \ms) \, r - V(\pi))^2] + \mE_{\calD}[(w_{\theta}(\mx, \phi_{\theta}(\ms)) \, r - V(\pi))^2] \\
    &= \mE_{p(\mx)\pi_0(\ms \,|\, \mx)p(r|\mx,\ms)} [ (w(\mx, \ms) - w(\ms, \phi_{\theta}(\ms)))^2 \, r^2 ] \\
    &= \mE_{p(\mx)\pi_0(\ms \,|\, \mx)} [ (w(\mx, \ms) - w(\ms, \phi_{\theta}(\ms)))^2 \, \mE_{p(r \,|\, \mx,\phi_{\theta}(\ms))} [r^2]] \\
    &= \mE_{p(\mx)} \Bigl[ \sum_{\ms \in \calS} \pi_0(\ms \,|\, \mx) \, (w(\mx, \ms) - w(\ms, \phi_{\theta}(\ms)))^2 \, \mE_{p(r \,|\, \mx,\phi_{\theta}(\ms))} [r^2] \Bigr] \\
    &= \mE_{p(\mx)} \Bigl[ \sum_{\ms \in \calS} \pi_0(\phi_{\theta}(\ms) \,|\, \mx) \frac{\pi_0(\ms \,|\, \mx)}{\pi_0(\phi_{\theta}(\ms) \,|\, \mx)} \, (w(\mx, \ms) - w(\ms, \phi_{\theta}(\ms)))^2 \, \mE_{p(r \,|\, \mx,\phi_{\theta}(\ms))} [r^2] \Bigr] \\
    &= \mE_{p(\mx)} \Bigl[ \sum_{z \in \calZ} \pi_0(z \,|\, \mx) \sum_{\ms \in \{\ms' \in\calS \mid \phi_{\theta}(\ms') = z\}} \pi_0(\ms \,|\, \mx, z)  (w(\mx, \ms) - w(\ms, z))^2 \, \mE_{p(r \,|\, \mx, z)} [r^2] \Bigr] \\
    &= \mE_{p(\mx)} \Bigl[ \sum_{z \in \calZ} \pi_0(z \,|\, \mx) \sum_{\ms \in \{\ms' \in\calS \mid \phi_{\theta}(\ms') = z\}} \pi_0(\ms \,|\, \mx, z) \\
    & \quad \quad \quad \quad \quad \quad \cdot \Bigl(w(\mx, \ms) - \sum_{\ms \in \{\ms' \in\calS \mid \phi_{\theta}(\ms') = z\}} \pi_0(\ms \,|\, \mx, z) w(\mx, \ms) \Bigr)^2 \, \mE_{p(r \,|\, \mx, z)} [r^2] \Bigr] \\
    &= \mE_{p(\mx)\pi_0(\phi_{\theta}(\ms) \,|\, \mx)} [ \mV_{\pi_0(\ms \,|\, \mx, \phi_{\theta}(\ms))}(w(\mx, \ms)) \, \mE_{p(r \,|\, \mx, z)} [r^2] ]
\end{align*}
\end{proof}

\subsection{The gain in MSE of LIPS with a stochastic slate abstraction}

In summary, LIPS has the following MSE gain over IPS with a stochastic slate abstraction.
\begin{align*}
    & n (\mathrm{MSE}(\ips) - \mathrm{MSE}(\ours)) \\
    &= \mE_{p(\mx)p_{\theta}(z\,|\,\mx,\pi_0)} \bigl[ \mV_{p_{\theta}(\ms\,|\,\mx,z,\pi_0)}(w(\mx, \ms)) \cdot \mE_{p_{\theta}(\ms\,|\,\mx,z,\pi_0)}\left[\mE_{p(r\,|\,\mx,\ms)}[r^2]\right] \bigr] \\
    &\quad + \mE_{p(\mx)p_{\theta}(z\,|\,\mx,\pi_0)} \bigl[ \mathrm{Cov}_{p_{\theta}(\ms\,|\,\mx,z,\pi_0)} \left( w(\mx, \ms)^2, \, \mE_{p(r\,|\,\mx, \ms)}[r^2] \right) \bigr] \\
    &\quad + 2 V(\pi) \mathrm{Bias}(\ours) + (1 - n) \mathrm{Bias}(\ours)^2
\end{align*}

\noindent Below, we provide the detailed deviation.
\begin{align*}
    & n ( \mathrm{MSE}(\ips) - \mathrm{MSE}(\ours) ) \\
    &= n( \mV_{\calD}(\ips) - \mV_{\calD}(\ours) - \mathrm{Bias}(\ours)^2 ) \\
    &= \mV_{\calD}(\ips) - \mV_{\calD}(\ours) - n \mathrm{Bias}(\ours)^2 \\
    &= \mE_{\calD}[(\ips - \mE_{\calD}[\ips])^2] - \mE_{\calD}[(\ours - \mE_{\calD}[\ours])^2 \\
    & \quad - n \mathrm{Bias}(\ours)^2 \\
    &= \mE_{\calD}[(\ips)^2] - (\mE_{\calD}[\ips])^2 - (\mE_{\calD}[(\ours)^2] - (\mE_{\calD}[\ours])^2 ) \\
    & \quad - n \mathrm{Bias}(\ours)^2 \\
    &= \mE_{\calD}[(\ips)^2] - \mE_{\calD}[(\ours)^2] - ((\mE_{\calD}[\ips])^2 - \mE_{\calD}[\ours])^2) \\
    & \quad - n \mathrm{Bias}(\ours)^2 \\
    &= \mE_{p(\mx)\pi_0(\ms\,|\,\mx)p_{\theta}(z\,|\,\mx,\ms)p(r\,|\,\mx,\ms)}[(w(\mx, \ms) r)^2 - (w_{\theta}(\mx, z) r)^2 ] \\
    & \quad - ((V(\pi))^2 - (V(\pi) + \mathrm{Bias}(\ours))^2) 
    - n \mathrm{Bias}(\ours)^2  \\
    &= \mE_{p(\mx)\pi_0(\ms\,|\,\mx)p_{\theta}(z\,|\,\mx,\ms)p(r\,|\,\mx,\ms)}[((w(\mx, \ms))^2 - (w_{\theta}(\mx, z))^2) \cdot r^2 ] \\
    & \quad + 2V(\pi) \mathrm{Bias}(\ours) + \mathrm{Bias}(\ours)^2 
    - n \mathrm{Bias}(\ours)^2  \\
    &= \mE_{p(\mx)\pi_0(\ms\,|\,\mx)p_{\theta}(z\,|\,\mx,\ms)}[(w(\mx, \ms)^2 - w_{\theta}(\mx, z)^2) \cdot \mE_{p(r\,|\,\mx,\ms)}[r^2] ] \\
    & \quad + 2 V(\pi) \mathrm{Bias}(\ours) + (1 - n) \mathrm{Bias}(\ours)^2
\end{align*}

\noindent Then, the first term is further decomposed into the variance and co-variance as follows.
\begin{align*}
    & \mE_{p(\mx)\pi_0(\ms\,|\,\mx)p_{\theta}(z\,|\,\mx,\ms)}[(w(\mx, \ms)^2 - w_{\theta}(\mx, z)^2) \cdot \mE_{p(r\,|\,\mx,\ms)}[r^2] ] \\
    &= \mE_{p(\mx)\pi_0(\ms\,|\,\mx)p_{\theta}(z\,|\,\mx,\ms)}[(w(\mx, \ms)^2 - w_{\theta}(\mx, z)^2) \cdot \mE_{p(r\,|\,\mx,\ms)}[r^2] ] \\
    &= \mE_{p(\mx)}[\sum_{\ms \in \calS} \pi_0(\ms\,|\,\mx) \sum_{z \in \calZ} p_{\theta}(z\,|\,\mx,\ms) (w(\mx, \ms)^2 - w_{\theta}(\mx, z)^2) \cdot \mE_{p(r\,|\,\mx,\ms)}[r^2] ] \\
    &= \mE_{p(\mx)}[\sum_{\ms \in \calS} \pi_0(\ms\,|\,\mx) \sum_{z \in \calZ} \frac{p_{\theta}(z\,|\, \mx,\pi_0) p_{\theta}(\ms\,|\,\mx, z, \pi_0)}{\pi_0(\ms\,|\,\mx)} (w(\mx, \ms)^2 - w_{\theta}(\mx, z)^2) \cdot \mE_{p(r\,|\,\mx,\ms)}[r^2] ] \\
    &= \mE_{p(\mx)}[\sum_{z \in \calZ} p_{\theta}(z\,|\, \mx, \pi_0) \sum_{\ms \in \calS} p_{\theta}(\ms\,|\,\mx, z, \pi_0) (w(\mx, \ms)^2 - w_{\theta}(\mx, z)^2) \cdot \mE_{p(r\,|\,\mx,\ms)}[r^2] ] \\
    &= \mE_{p(\mx)p_{\theta}(z\,|\,\mx,\pi_0)p_{\theta}(\ms\,|\,\mx,z,\pi_0)}[(w(\mx, \ms)^2 - w_{\theta}(\mx, z)^2) \cdot \mE_{p(r\,|\,\mx,\ms)}[r^2] ] \\
    &= \mE_{p(\mx)p_{\theta}(z\,|\,\mx,\pi_0)p_{\theta}(\ms\,|\,\mx,z,\pi_0)}[(w(\mx, \ms)^2 - (\mE_{p_{\theta}(\ms\,|\,\mx,z,\pi_0)}[w(\mx, \ms)])^2) \cdot \mE_{p(r\,|\,\mx,\ms)}[r^2] ] \\
    &= \mE_{p(\mx)p_{\theta}(z\,|\,\mx,\pi_0)p_{\theta}(\ms\,|\,\mx,z,\pi_0)} [ ( w(\mx, \ms)^2 - ( \mE_{p_{\theta}(\ms\,|\,\mx,z,\pi_0)}[w(\mx,\ms)] )^2 ) \\
    & \quad \quad \quad \cdot ( \mE_{p_{\theta}(\ms\,|\,\mx,z,\pi_0)}[\mE_{p(r\,|\,\mx,\ms)}[r^2]] + ( \mE_{p(r\,|\,\mx, \ms)}[r^2] - \mE_{p_{\theta}(\ms\,|\,\mx,z,\pi_0)}[\mE_{p(r\,|\,\mx,\ms)}[r^2]] ) ) ] \\
    &= \mE_{p(\mx)p_{\theta}(z\,|\,\mx,\pi_0)p_{\theta}(\ms\,|\,\mx,z,\pi_0)} [ ( w(\mx, \ms)^2 - ( \mE_{p_{\theta}(\ms\,|\,\mx,z,\pi_0)}[w(\mx,\ms)] )^2 ) \cdot \mE_{p_{\theta}(\ms\,|\,\mx,z,\pi_0)}[\mE_{p(r\,|\,\mx,\ms)}[r^2]] ] \\
    & \quad + \mE_{p(\mx)p_{\theta}(z\,|\,\mx,\pi_0)p_{\theta}(\ms\,|\,\mx,z,\pi_0)} [ ( w(\mx, \ms)^2 - ( \mE_{p_{\theta}(\ms\,|\,\mx,z,\pi_0)}[w(\mx,\ms)] )^2 ) \\
    & \quad \quad \quad \cdot ( \mE_{p(r\,|\,\mx, \ms)}[r^2] - \mE_{p_{\theta}(\ms\,|\,\mx,z,\pi_0)}[\mE_{p(r\,|\,\mx,\ms)}[r^2]] ) ] \\
    &= \mE_{p(\mx)p_{\theta}(z\,|\,\mx,\pi_0)} \bigl[ \mV_{p_{\theta}(\ms\,|\,\mx,z,\pi_0)}(w(\mx, \ms)) \cdot \mE_{p_{\theta}(\ms\,|\,\mx,z,\pi_0)}\left[\mE_{p(r\,|\,\mx,\ms)}[r^2]\right] \bigr] \\
    & \quad + \mE_{p(\mx)p_{\theta}(z\,|\,\mx,\pi_0)} \bigl[ \mathrm{Cov}_{p_{\theta}(\ms\,|\,\mx,z,\pi_0)} \left( w(\mx, \ms)^2, \, \mE_{p(r\,|\,\mx, \ms)}[r^2] \right) \bigr]
\end{align*}

\end{document}